\documentclass[10pt,journal,compsoc]{IEEEtran}
\usepackage{tikz}
\usetikzlibrary{positioning, automata, chains, arrows.meta}
\usetikzlibrary{matrix,positioning,calc}
\usepackage{booktabs}
\usepackage[utf8]{inputenc}
\usepackage{geometry}
\usepackage{graphicx}
\usepackage{multirow}
\usepackage{amsmath}
\usepackage{amsthm}
\usepackage{stackengine}
\usepackage{float}

\usepackage{amssymb}
\usepackage{algorithm}
\usepackage{algorithmicx}
\usepackage{algpseudocode} 
\usepackage{makecell}
\usepackage{color}
\usepackage{tikz}
\usepackage{caption}
\usepackage{lscape}
\usepackage{pdflscape}
\usepackage{rotating}
\usepackage{empheq}
\usepackage{longtable}
\usepackage{MnSymbol}%
\usepackage{wasysym}

\usepackage{url}
\usepackage{hyperref}  
\usepackage{color}

\hypersetup{
    colorlinks=true,
    linkcolor=blue,
    filecolor=blue,      
    urlcolor=blue,
    citecolor=blue
}
\usetikzlibrary{positioning, automata, chains, arrows.meta}
\usetikzlibrary{matrix,positioning,calc}

\newcommand{\True}{\mbox{{1}}}
\newcommand{\False}{\mbox{{0}}}

\newtheorem{mytheorem}{Theorem}
\newtheorem{mylemma}{Lemma}

\newtheorem{remark}{Remark}

\algnewcommand{\LineComment}[1]{\State \(\triangleright\) #1}

\ifCLASSOPTIONcompsoc
  \usepackage[nocompress]{cite}
\else
  \usepackage{cite}
\fi

\ifCLASSINFOpdf
 
\else
 
\fi

\begin{document}

\title{On the Convergence of Tsetlin Machines for the IDENTITY- and NOT Operators}

\author{Xuan~Zhang,~Lei~Jiao,~\IEEEmembership{Senior member,~IEEE,}
        Ole-Christoffer~Granmo,~and~Morten~Goodwin
\IEEEcompsocitemizethanks{\IEEEcompsocthanksitem X. Zhang is with the Centre for Artificial Intelligence Research, University of Agder, Grimstad, Norway, and with the Norwegian Research Centre (NORCE), Grimstad, Norway, and also with the Confirmit AS, Norway.\protect
\IEEEcompsocthanksitem L. Jiao, O.-C. Granmo, and M. Goodwin are with the Centre for Artificial Intelligence Research, University of Agder, and the Department of ICT, University of Agder, Grimstad, Norway.
\IEEEcompsocthanksitem Corresponding author: Lei Jiao. E-mail: lei.jiao@uia.no.
}
}

%

\IEEEtitleabstractindextext{%
\begin{abstract}
The Tsetlin Machine (TM) is a recent machine learning algorithm with several distinct properties, such as interpretability, simplicity, and hardware-friendliness. Although numerous empirical evaluations report on its performance, the mathematical analysis of its convergence is still open. In this article, we analyze the convergence of the TM with only one clause involved for classification. More specifically, we examine two basic logical operators, namely, the “IDENTITY”- and “NOT” operators. Our analysis reveals that the TM, with just one clause, can converge correctly to the intended logical operator, learning from training data over an infinite time horizon. Besides, it can capture arbitrarily rare patterns and select the most accurate one when two candidate patterns are incompatible, by configuring a granularity parameter. The analysis of the convergence of the two basic operators lays the foundation for analyzing other logical operators. These analyses altogether, from a mathematical perspective, provide new insights on why TMs have obtained state-of-the-art performance on several pattern recognition problems.
\end{abstract}

}

\maketitle

\IEEEdisplaynontitleabstractindextext

\IEEEpeerreviewmaketitle

\IEEEraisesectionheading{\section{Introduction}\label{sec:introduction}}
The Tsetlin Machine (TM) \cite{granmo2018tsetlin} is a novel machine learning mechanism in which groups of Tsetlin Automata (TAs)  \cite{Tsetlin1961} operate on binary data using propositional logic, to learn patterns from training samples.   A game between the TAs organizes the learning in order to optimize pattern classification accuracy. The TM is an offspring of learning automata (LAs) that have been studied for decades and employed in many applications~\cite{sarkar2000supervised,jiao2016optimizing, yang2020learning}. The state-of-the-art of LA research is surveyed in \cite{zhang2019conclusive, yazidi2019hierarchical}. 

The main advantages of TMs are twofold. The first advantage is their transparency, i.e., the ability to lay open the reasoning behind the decision making process. This advantage addresses a critical challenge in Artificial Intelligence (AI) research -- lack of interpretability~\cite{ribeiro2016should}. Deep neural networks, in particular, mainly support \emph{approximate} local interpretability through post-processing. The approximations do not guarantee model fidelity, however, and global interpretability is currently out of reach~\cite{Rudin2019}. TMs, on the other hand, is inherently interpretable, being founded on propositional logic. That is, one can formulate and manipulate patterns using logical operators. As a result, there exist closed formula expressions for both local and global TM interpretation, akin to SHAP \cite{blakely2021closed}.

The second advantage of TMs is simplicity and hardware-friendliness \cite{wheeldon2020learning}. Because each  learning component of a TM is a finite-state automaton (i.e., the TA), the TM only needs to maintain a set of integers to track its state. In contrast to the more extensive arithmetic operations required by most AI techniques, a TA learns using increment and decrement operations only. TMs are thus naturally suited for hardware implementation. Indeed, due to the robustness of TA learning and TM pattern representation, TMs have recently been shown to be inherently fault-tolerant, completely masking stuck-at faults~\cite{shafik2020explainability}.


{\color{black}
The TM itself can further act as a building block in more advanced architectures. It has for instance been extended to support convolution \cite{granmo2019convolutional}, regression~\cite{abeyrathna2019nonlinear, abeyrathna2020integerregression}, and relational reasoning \cite{saha2021relational}. Computation-wise, the native parallelism of TMs supports efficient deployment on Compute Unified Device Architecture (CUDA) \cite{abeyrathna2020massively}. The various TM architectures have been employed in several application domains with promising results. One example is natural language processing where the propositional TM formulas capture textual patterns to solve tasks such as text classification \cite{berge2019text,yadav2021dwr,nicolae2021question}, word sense disambiguation \cite{yadav2021wordsense}, novelty detection~\cite{bhattarai2021novelty}, semantic relation analysis \cite{saha2020causal}, question-answering \cite{saha2021relational}, keyword spotting \cite{lei2021kws}, and aspect-based sentiment analysis \cite{rohan2021AAAI} . These studies, overall, report that TMs, with smaller memory footprint and higher computational efficiency, obtain better or competitive classification accuracy compared to state-of-the-art AI techniques. Further, the applications demonstrate how the TM facilitates interpretation, contrasted against other machine learning techniques.

}

Although results from computational simulations and empirical evaluations have demonstrated the effectiveness of TMs, analyzing the convergence of the TM learning process is yet to be done. We establish the first step in this undertaking in the present article by analyzing the convergence of TMs via rigorous mathematical tools.  The proofs employ a quasi-stationary analysis of the system states together with Markov chain analysis. Overall, we aim to analyze four fundamental logical operators: IDENTITY/NOT, AND, OR, and XOR. The analysis of these operators forms the foundation for analyzing more complex propositional expressions.  In this article, we study the convergence of the IDENTITY- and NOT operators, while we will address the AND-, OR-, and XOR- operators in separate articles.

The IDENTITY- and NOT operators are unary, operating on  one bit of data. We refer to the analysis of these operators as the ``one-bit'' case. The ``two-bit'' case deals with the binary operators AND, OR, and XOR. For the one-bit case, we first prove that the TM can converge surely to the correct pattern when the training data is noise-free. Then we generalize the one-bit case by analyzing the effect of noise, establishing how the noise probability of the data and the granularity parameter of the TM govern convergence.

The remainder of the paper is organized as follows. Section \ref{Review} briefly reviews the TM and its training process for the one-bit case. In Section~\ref{sect:one-bitcase}, we present our analytical procedure for the one-bit case and the main analytical results. We conclude the paper in Section \ref{conclusion}.

\section{Review of the Tsetlin Machine}\label{Review}
The fundamentals of TMs have been detailed in \cite{granmo2018tsetlin}. To make this article self-contained, in this section, we do a brief review on TMs, which includes the concept of TA, the TM architecture, and the training process of the TM for the one-bit case.

\subsection{Tsetlin Automata (TAs)}\label{TA}
TAs are the basic components of the TM. A TA is a fixed structure deterministic automaton that performs actions in a stochastic environment. By interacting with the environment, it aims to learn the optimal action, i.e., the one that has the highest probability of eliciting a reward. A complete description of TA can be found in~\cite{Tsetlin1961,Narendra1989LearningIntroduction}. Figure~\ref{figure:TAarchitecture_basic} shows a simple example of a two-action TA with $2N$ memory states, where $N\in[1, +\infty)$. When the system is in state 1 to $N$, i.e., on the left-hand side of the state-space, the TA chooses Action 1. If the system is in state $N+1$ to $2N$, i.e., on the right-hand side of the state-space, the action of the TA is Action 2.

In each interaction with the environment, the TA performs one of the available actions. The environment, in turn, responds with either a reward or a penalty. If the TA receives a penalty, it will move towards the opposite side of the current action. The upper state chain shows this in Figure~\ref{figure:TAarchitecture_basic}. Conversely, if the TA receives a reward as a response, it will switch to a ``deeper'' state by moving towards the left or right end of the state chain, depending on whether the current action is Action 1 or Action 2. The lower state chain shows this in Figure~\ref{figure:TAarchitecture_basic}.

Note that the number of states in a TA can be adjusted. The larger the number, the slower the TA converges, however, the more accurately the TA performs in the stochastic environment. The goal of the TA is to converge to the optimal action by learning. For example, if Action 1 provides a reward with probability 90\% and Action 2 has a reward probability of 80\%, we expect the TA only to perform Action 1 after the training process.

\begin{figure}
\centering
\begin{minipage}{1\textwidth}
\resizebox{0.5\textwidth}{!}{
\begin{tikzpicture}[node distance = .35cm, font=\Huge]
    \tikzstyle{every node}=[scale=0.35]
    \node[state] (A) at (0,2) {~~~1~~~~};
    \node[state] (B) at (1.5,2) {~~~2~~~~};
    
    \node[state,draw=white] (M) at (3,2) {~~~$....$~~~};
    
    \node[state] (C) at (4.5,2) {$N-1$};
    \node[state] (D) at (6,2) {~~~\!$N$~~~};
    
    \node[state] (E) at (7.5,2) {$N+1$};
    \node[state] (F) at (9,2) {$N+2$};
    
    \node[state,draw=white] (G) at (10.5,2) {~~~$....$~~~};
    
    \node[state] (H) at (12,2) {\!$2N-1$};
    \node[state] (I) at (13.5,2) {~~\!$2N$~~~~};
    
       \node[state] (J) at (0,0) {~~~1~~~~};
    \node[state] (K) at (1.5,0) {~~~2~~~~};
    
    \node[state,draw=white] (L) at (3,0) {~~~$....$~~~};
    
    \node[state] (N) at (4.5,0) {$N-1$};
    \node[state] (O) at (6,0) {~~~\!$N$~~~};
    
    \node[state] (P) at (7.5,0) {$N+1$};
    \node[state] (Q) at (9,0) {$N+2$};
    
    \node[state,draw=white] (R) at (10.5,0) {~~~$....$~~~};
    
    \node[state] (S) at (12,0) {\!$2N-1$};
    \node[state] (T) at (13.5,0) {~~\!$2N$~~~~};

    \node[thick] at (4,4) {$Action~1$};
    \node[thick] at (9.5,4) {$Action~2$};
    
    \node[thick] at (0,1) {$Reward~(R):$};
    \node[thick] at (0,3) {$Penalty~(P):$};

    \draw[every loop]
    (A) edge[bend left] node [scale=1.2, above=0.1 of B]{} (B)
    (B) edge[bend left] node  [scale=1.2, above=0.1 of M] {} (M)
    (M) edge[bend left] node  [scale=1.2, above=0.1 of C] {} (C)
    (C) edge[bend left] node [scale=1.2, above=0.1 of D] {} (D)
    (D) edge[bend left] node  [scale=1.2, above=0.1 of E] {} (E);

    \draw[every loop]
    (I) edge[bend left] node [scale=1.2, below=0.1 of H] {} (H)
    (H) edge[bend left] node  [scale=1.2, below=0.1 of G] {} (G)
    (G) edge[bend left] node [scale=1.2, below=0.1 of F] {} (F)
    (F) edge[bend left] node  [scale=1.2, below=0.1 of E] {} (E)
    (E) edge[bend left] node  [scale=1.2, below=0.1 of D] {} (D);

    
    \draw[every loop]
    (P) edge[bend left] node  [scale=1.2, above=0.1 of P] {} (Q)
    (Q) edge[bend left] node [scale=1.2, above=0.1 of Q] {} (R)
    (R) edge[bend left] node [scale=1.2, above=0.1 of R] {} (S)
    (S) edge[bend left] node [scale=1.2, above=0.1 of S] {} (T)
    (T) edge[loop right] node [scale=1.2, below=0.1 of T] {} (T);
    
    \draw[every loop]
 
    (O) edge[bend left ] node [scale=1.2, below=0.1 of N] {} (N)
    (N) edge[bend left] node  [scale=1.2, below=0.1 of L] {} (L)
    (L) edge[bend left] node  [scale=1.2, below=0.1 of K] {} (K)
    (K) edge[bend left ] node [scale=1.2, below=0.1 of J] {} (J)
    (J) edge[loop left] node [scale=1.2, below=0.1 of J] {} (J);
    
      \draw[dotted, thick] (6.75,-0.6) -- (6.75,3);

\end{tikzpicture}
}
\end{minipage}
\caption{A two-action Tsetlin Automaton with $2N$ states.}
\label{figure:TAarchitecture_basic}
\end{figure}
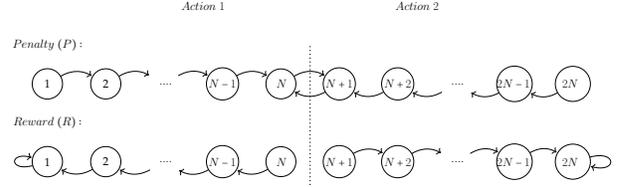
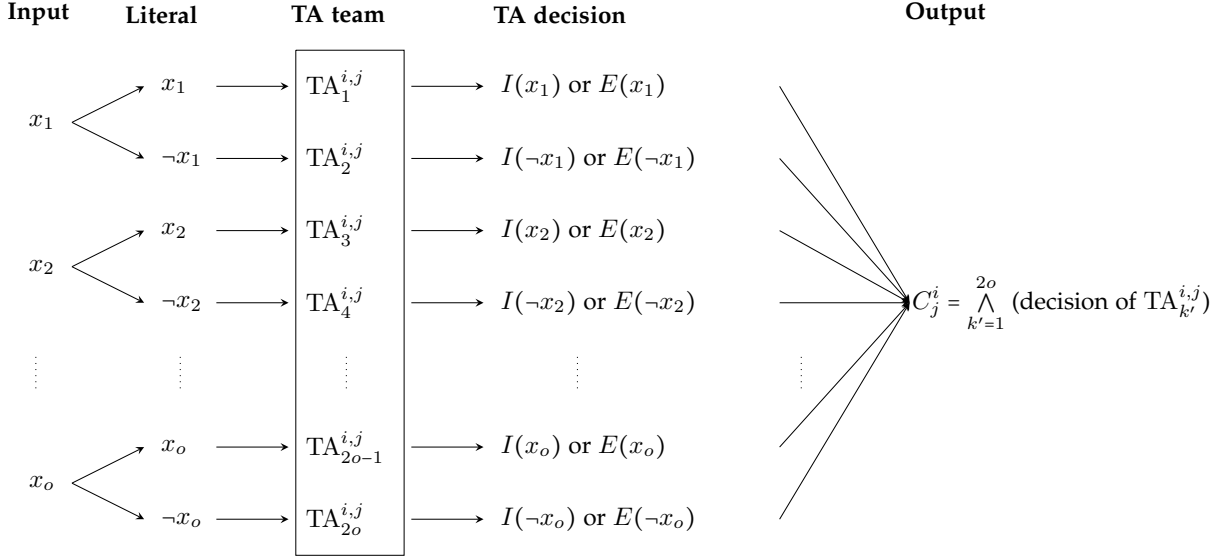
\begin{figure*}[htbp]
\begin{center}
\begin{minipage}{1\textwidth}
\resizebox{1\textwidth}{!}{
\begin{tikzpicture}[node distance = .35cm]

    \node[label=left:{\bf Input}] at (0.2,1.5) {};
    \node[label=right:{\bf Literal}] at (0.5,1.5) {};
    \node[label=right:{\bf TA team}] at (2.8,1.5) {};
    \node[label=right:{\bf TA decision}] at (5.6,1.5) {};
    \node[label=right:{\bf Output}] at (11.3,1.5) {};

    \node[label=left:$x_1$] at (0,0) {};
    \node[label=left:$x_2$] at (0,-2) {};
    \node[label=left:$x_o$] at (0,-5) {};
    
    \node[label=right:$x_1$] at (1,0.5) {};
    \node[label=right:$\neg x_1$] at (1,-0.5) {};
    \node[label=right:$x_2$] at (1,-1.5) {};
    \node[label=right:$\neg x_2$] at (1,-2.5) {};
    \node[label=right:$x_o$] at (1,-4.5) {};
    \node[label=right:$\neg x_o$] at (1,-5.5) {};
    
    \node[label=right:$\mathrm{TA}_1^{i,j}$] at (3,0.5) {};
    \node[label=right:$\mathrm{TA}_2^{i,j}$] at (3,-0.5) {};
    \node[label=right:$\mathrm{TA}_3^{i,j}$] at (3,-1.5) {};
    \node[label=right:$\mathrm{TA}_4^{i,j}$] at (3,-2.5) {};
    \node[label=right:$\mathrm{TA}_{2o-1}^{i,j}$] at (3,-4.5) {};
    \node[label=right:$\mathrm{TA}_{2o}^{i,j}$] at (3,-5.5) {};
    \draw (3.1, 1) -- (4.6, 1) -- (4.6, -6) -- (3.1, -6) -- (3.1, 1); 
    
    \node[label=right:$I(x_1)~\text{or}~ E(x_1)$] at (5.7,0.5) {};
    \node[label=right:$I(\neg x_1)~\text{or}~ E(\neg x_1)$] at (5.7,-0.5) {};
    \node[label=right:$I(x_2)~\text{or}~ E(x_2)$] at (5.7,-1.5) {};
    \node[label=right:$I(\neg x_2)~\text{or}~ E(\neg x_2)$] at (5.7,-2.5) {};
    \node[label=right:$I(x_o)~\text{or}~ E(x_o)$] at (5.7,-4.5) {};
    \node[label=right:$I(\neg x_o)~\text{or}~ E(\neg x_o)$] at (5.7,-5.5) {};
    
    \node[label=right:{$C^i_j= \bigwedge\limits_{k'=1}^{2o}$ (decision of $\mathrm{TA}_{k'}^{i,j}$)}] at (11.4,-2.5) {};
    
    \draw [-{stealth[length=4mm]}] (0,0) -- (1,0.5);
    \draw [-{stealth[length=4mm]}] (0,0) -- (1,-0.5);
    \draw [-{stealth[length=4mm]}] (0,-2) -- (1,-1.5);
    \draw [-{stealth[length=4mm]}] (0,-2) -- (1,-2.5);
    \draw [-{stealth[length=4mm]}] (0,-5) -- (1,-4.5);
    \draw [-{stealth[length=4mm]}] (0,-5) -- (1,-5.5);
    
    \draw [-{stealth[length=4mm]}] (2,0.5) -- (3, 0.5);
    \draw [-{stealth[length=4mm]}] (2,-0.5) -- (3, -0.5);
    \draw [-{stealth[length=4mm]}] (2,-1.5) -- (3, -1.5);
    \draw [-{stealth[length=4mm]}] (2,-2.5) -- (3, -2.5);
    \draw [-{stealth[length=4mm]}] (2,-4.5) -- (3, -4.5);
    \draw [-{stealth[length=4mm]}] (2,-5.5) -- (3, -5.5);
    
    \draw [-{stealth[length=4mm]}] (4.7,0.5) -- (5.7, 0.5);
    \draw [-{stealth[length=4mm]}] (4.7,-0.5) -- (5.7, -0.5);
    \draw [-{stealth[length=4mm]}] (4.7,-1.5) -- (5.7, -1.5);
    \draw [-{stealth[length=4mm]}] (4.7,-2.5) -- (5.7, -2.5);
    \draw [-{stealth[length=4mm]}] (4.7,-4.5) -- (5.7, -4.5);
    \draw [-{stealth[length=4mm]}] (4.7,-5.5) -- (5.7, -5.5);
    
    \draw [-{stealth[length=4mm]}] (9.8,0.5) -- (11.6, -2.5);
    \draw [-{stealth[length=4mm]}] (9.8,-0.5) -- (11.6, -2.5);
    \draw [-{stealth[length=4mm]}] (9.8,-1.5) -- (11.6, -2.5);
    \draw [-{stealth[length=4mm]}] (9.8,-2.5) -- (11.6, -2.5);
    \draw [-{stealth[length=4mm]}] (9.8,-4.5) -- (11.6, -2.5);
    \draw [-{stealth[length=4mm]}] (9.8,-5.5) -- (11.6, -2.5);

    \draw
    [dotted] (-0.5,-3.25) -- (-0.5,-3.75)
    [dotted] (1.5,-3.25) -- (1.5,-3.75)
    [dotted] (3.8,-3.25) -- (3.8,-3.75)
    [dotted] (7,-3.25) -- (7,-3.75)
    [dotted] (10.1,-3.25) -- (10.1,-3.75);
    
\end{tikzpicture}
}

\end{minipage}
\end{center}
\caption{\label{fig:tateam} A TA team $G^i_j$ consisting of $2o$ TAs.}
\end{figure*}
\begin{figure}[htbp]
\begin{center}
\begin{minipage}{1\textwidth}
\resizebox{0.4\textwidth}{!}{
\begin{tikzpicture}[node distance = .35cm]

\node[label=left:TA team $1$] at (0,0) {};
\node[label=left:TA team $2$] at (0,-1) {};
\node[label=left:TA team $m$] at (0,-4) {};
\draw (-2.3, 0.4) -- (-2.3, -0.4) -- (0, -0.4) -- (0, 0.4) -- (-2.3, 0.4);
\draw (-2.3, -0.6) -- (-2.3, -1.4) -- (0, -1.4) -- (0, -0.6) -- (-2.3, -0.6);
\draw (-2.3, -3.6) -- (-2.3, -4.4) -- (0, -4.4) -- (0, -3.6) -- (-2.3, -3.6);

\node[label=right: $C^i_1$] at (1,0) {};
\node[label=right: $C^i_2$] at (1,-1) {};
\node[label=right: $C^i_m$] at (1,-4) {};

\node[label=right: $\bigvee\limits_{j=1}^{m}C^i_j$] at (3.5,-2) {};
    
\draw [-{stealth[length=4mm]}] (0.1,0) -- (1,0);
\draw [-{stealth[length=4mm]}] (0.1,-1) -- (1,-1);
\draw [-{stealth[length=4mm]}] (0.1,-4) -- (1,-4);

\draw [-{stealth[length=4mm]}] (2,0) -- (3.5, -2);
\draw [-{stealth[length=4mm]}] (2,-1) -- (3.5, -2);
\draw [-{stealth[length=4mm]}] (2,-4) -- (3.5, -2);
    
\draw [dotted] (-1.5,-2.2) -- (-1.5,-2.8);
\draw [dotted] (1.5,-2.2) -- (1.5,-2.8);
\draw [dotted] (2.3,-2.2) -- (2.3,-2.8);

\end{tikzpicture}
}

\end{minipage}
\end{center}
\caption{\label{fig:TMDisjunction} TM disjunctive normal form architecture.}
\end{figure}
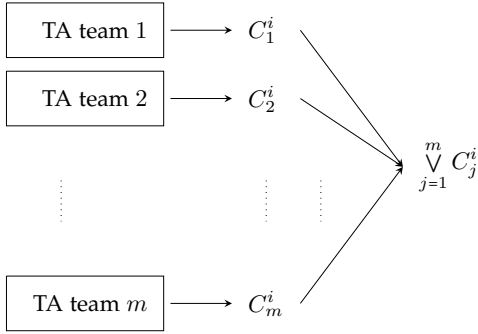
\begin{figure}[htbp]
\begin{center}
\begin{minipage}{1\textwidth}
\resizebox{0.5\textwidth}{!}{
\begin{tikzpicture}[node distance = .35cm]

\node[label=left:TA team $1~~~~~~$] at (0,0) {};
\node[label=left:TA team $2~~~~~~$] at (0,-1) {};
\node[label=left:TA team $m-1$] at (0,-3) {};
\node[label=left:TA team $m~~~~$] at (0,-4) {};
\draw (-3, 0.4) -- (-3, -0.4) -- (0, -0.4) -- (0, 0.4) -- (-3, 0.4);
\draw (-3, -0.6) -- (-3, -1.4) -- (0, -1.4) -- (0, -0.6) -- (-3, -0.6);
\draw (-3, -2.6) -- (-3, -3.4) -- (0, -3.4) -- (0, -2.6) -- (-3, -2.6);
\draw (-3, -3.6) -- (-3, -4.4) -- (0, -4.4) -- (0, -3.6) -- (-3, -3.6);

\node[label=right: $C^i_1$] at (1,0) {};
\node[label=right: $C^i_2$] at (1,-1) {};
\node[label=right: $C^i_{m-1}$] at (1,-3) {};
\node[label=right: $C^i_m$] at (1,-4) {};
\node[label=right:$+$] at (2,0) {};
\node[label=right:$-$] at (2,-1) {};
\node[label=right:$+$] at (2,-3) {};
\node[label=right:$-$] at (2,-4) {};

\node[label=right: $\sum\limits_{j=1}^{m}(-1)^{(j-1)} C^i_j$] at (3.5,-2) {};
    
\draw [-{stealth[length=4mm]}] (0.1,0) -- (1,0);
\draw [-{stealth[length=4mm]}] (0.1,-1) -- (1,-1);
\draw [-{stealth[length=4mm]}] (0.1,-3) -- (1,-3);
\draw [-{stealth[length=4mm]}] (0.1,-4) -- (1,-4);

\draw [-{stealth[length=4mm]}] (2.6,0) -- (3.5, -2);
\draw [-{stealth[length=4mm]}] (2.6,-1) -- (3.5, -2);
\draw [-{stealth[length=4mm]}] (2.6,-3) -- (3.5, -2);
\draw [-{stealth[length=4mm]}] (2.6,-4) -- (3.5, -2);
    
\draw [dotted] (-2.5,-1.8) -- (-2.5,-2.2);
\draw [dotted] (1.5,-1.8) -- (1.5,-2.2);
\draw [dotted] (3,-1.8) -- (3,-2.2);

\end{tikzpicture}
}

\end{minipage}
\end{center}
\caption{\label{fig:TMVoting} TM voting architecture.}
\end{figure}
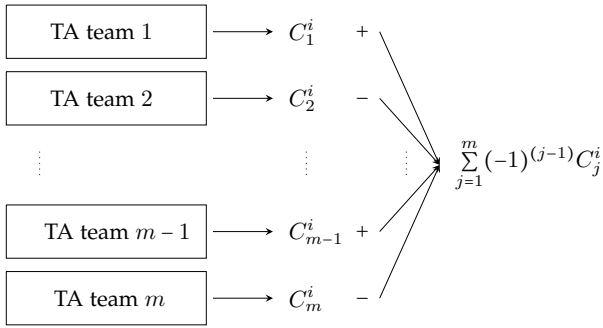

\subsection{Tsetlin Machines (TMs)}\label{sect:TM}
{\bf Tsetlin Automata Team.} {\color{black}A TM consists of $m$ teams of TAs that participate in a decentralized game, designed to solve pattern recognition tasks. The TM trains the TA teams to produce $m$ conjunctive clauses $C^i_j,~j=1,2,...,m$, defined below. These capture the sub-patterns that characterize a particular class $i$. } 

The TM takes binary inputs and utilizes propositional logic to represent patterns. In the general case, suppose $X=[x_1, x_2, \ldots, x_o]$, with $x_k \in \{0, 1\}, k=1, 2, \ldots, o$, is the input of the TM. Then each TA team is formed by $o$ pairs of TAs, each of which takes care of one input value $x_k$. More specifically, a TA team $\mathcal{G}^i_j=\{\mathrm{TA}^{i,j}_{k'}|1\leq k'\leq 2o\}$ consists of $2o$ TAs, where $i$ refers to a specific pattern class, while $j$ refers to a specific clause. The automaton $\mathrm{TA}^{i,j}_{2k-1}$ addresses the input $x_k$ as is, whereas $\mathrm{TA}^{i,j}_{2k}$ addresses the negation of $x_k$, i.e., $\neg x_k$. Note that the inputs and their negations are jointly referred to as literals. Each TA, in turn, chooses one of two actions. It either ``includes'' or ``excludes'' its literal. 
{\color{black}The actions taken by the TA team members then decide whether the associated literals take part in a conjunction, expressed by the conjunctive clause~\cite{granmo2018tsetlin}:

{\footnotesize
\begin{equation}
\label{eqn:clause1}
C^i_j(X) = \begin{cases}
\left(\bigwedge\limits_{k \in I^i_j} {x_k}\right) \wedge \left(\bigwedge\limits_{k \in \bar{I}^i_j} {\neg x_k}\right) \wedge 1 & \mathrm{During\ training},\\
\left(\left(\bigwedge\limits_{k \in I^i_j} {x_k}\right) \wedge \left(\bigwedge\limits_{k \in \bar{I}^i_j} {\neg x_k}\right)\right) \vee 0 & \mathrm{During\ testing}.
\end{cases}
\end{equation}
}
\noindent Above, $I^i_j$ and $\bar{I}^i_j$ 
are the subsets of indexes corresponding to the literals that have been included in the clause. In the subset $I^i_j$, the included literals are the original input $x_k$, whereas in $\bar{I}^i_j$, 
the included literals are the negated input $\neg x_k$. A clause does not need to include literals, i.e., $I^i_j=\emptyset$ and $\bar{I}^i_j=\emptyset$ are allowed. This happens when all the TAs of the clause choose to exclude their literals. We then define the value of the clause to be $C^i_j(X)=1$ during the training process and $C^i_j(X)=0$ during the testing process. Note that an empty clause is usually defined to be  $1$-valued in propositional logic. However, empirically, it turns out that TMs in general obtain higher test accuracy (on new data) when the empty clauses are $0$-valued. In this paper, our theoretical analysis addresses the training process, which means that we operate with the standard definition of empty clauses. 

To illustrate the structure of a clause and its relationship to its literals, we show a general clause with $2o$ TAs in Figure \ref{fig:tateam}. For ease of notation, we define the include action as $I(x)=x$ and $~I(\neg x)=\neg x$, while we define the exclude action as $E(\cdot)=1$, the latter meaning that an excluded literal does not contribute to the conjunctive output.
}

{\bf Disjunctive Normal Form (DNF) Architecture.} Multiple TA teams are finally assembled into a complete TM. Figure \ref{fig:TMDisjunction} shows a TM composed of $m$ TA teams. The goal of the TM is to learn a pattern representation of a given class~$i$. To this end, the TM decomposes the class into multiple sub-patterns. The outputs of the TA teams, i.e., the conjunctive clauses $C^i_j, 1 \leq j \leq m$, are used to represent the sub-patterns of the class.  That is, the disjunction of the clause sub-patterns represents the class:
\begin{align}
\label{eqn:Phi}
\hat{y}^i = \bigvee\limits_{j=1}^{m}C^i_j.
\end{align}
This is equivalent to saying that if an input ``matches'' any one of the clauses, i.e., $C^i_j(X) = 1$, for any $j=1,2,\ldots, m$, then it belongs to class $i$. Accordingly, the output of the TM is $\hat{y}^i=1$. Otherwise, the output is $\hat{y}^i=0$.

{\bf Voting Architecture.} Besides the DNF architecture, one can organize the TA teams in a ``voting'' architecture. Figure \ref{fig:TMVoting} shows such a TM. In this architecture, the TM assigns a polarity to each TA team. TA teams with odd indexes get positive polarity, and they vote for class $i$. The remaining TA teams get negative polarity and vote against class $i$. The voting consists of summing the output of the clauses, according to polarity:
\begin{equation}
\label{eqn:summation}
    f(X)= \left( \sum\limits_{j \in \{1, 3, \ldots, m-1\}} C_j^i(X) \right) - \left(\sum\limits_{j \in \{2, 4, \ldots, m\}}  C_j^i(X) \right).
\end{equation}
The output of the TM, in turn, is decided by the unit step function:
\begin{align}
\label{eqn:yivoting}
\hat{y}^i={\begin{cases}\False&{\text{for }}f(X)<0\\\True&{\text{for }}f(X)\geq 0\end{cases}}.
\end{align} 
Compared with the DNF architecture, the voting architecture turns out to be more robust to noise and thus more suitable for solving real-world problems~\cite{granmo2018tsetlin}.

{\bf Remark.} One drawback of employing the decentralized TA teams for learning is that sub-patterns, in effect, must compete for clauses. It thus can happen that all the available clauses converge to one or some of the sub-patterns. The TM would thus fail to capture the other sub-patterns of the class. To address this drawback, the author of \cite{granmo2018tsetlin} introduced voting summation target $T$ to bound the maximum number of clauses assigned to each sub-pattern. By doing so, the TM distributes the clauses (i.e., its pattern representation capacity) among as many sub-patterns as possible.  The one-bit case does not require such a summation target. We, therefore, do not go into the details of this mechanism in the present paper.

\subsection{The TM Game for Learning Patterns}\label{sect:TMTraining}
\subsubsection{The TM Game}\label{sect:TMGame}

{\color{black}
The TM training process is built on interconnecting the TAs as players in a decentralized game, with the game being driven by the training data. The training data $(X=[x_1,x_2,...,x_o],~y^i)$ is obtained from a dataset $\mathcal{S}$, distributed according to the probability distribution $P(X, y^i)$.  In the game, each TA is guided by so-called Type I Feedback and Type II Feedback, as shown in Table \ref{table:type_i_feedback} and Table \ref{table:type_ii_feedback}, respectively. Type I Feedback is activated when the training sample fed into the system has a positive label, i.e., $y^i=1$, meaning that the sample belongs to class $i$. When the training sample is labeled as not belonging to class $i$, i.e., $y^i=0$, Type II Feedback is triggered for generating responses. These two types of feedback are designed to reinforce true positive output, i.e., $(\hat{y}^i=1, y^i=1)$. 
Simultaneously, they suppress false positive output, i.e., $(\hat{y}^i=1, y^i=0)$, and false negative output, i.e., $(\hat{y}^i=0, y^i=1)$.

As shown in Table \ref{table:type_i_feedback}, Type I Feedback encourages a TA to include its literal into its clause whenever the literal has value $1$ and the clause evaluates to $1$. This is achieved by assigning a relatively large probability $\frac{s-1}{s}$ for rewarding the inclusion of $1$-valued literals and for penalizing their exclusion, where $s \geq 1$ is the so called granularity parameter. Conversely, if the clause falsely evaluates to $0$, Type I Feedback leads the TA to exclude its literal by rewarding exclude and penalizing include, now with probability $\frac{1}{s}$. Accordingly, excluding a sufficient number of literals makes the clause eventually evaluate to $1$. When the forces of including literals and excluding literals are in balance, the TM reaches a Nash equilibrium. In this manner, Type I Feedback reinforces the true positive output, and suppresses the false negative output.

Type II Feedback, shown in Table \ref{table:type_ii_feedback}, penalizes excluding literals of value $0$ with probability $1.0$, however, only when the clause falsely evaluates to $1$. This is to encourage including literals of value $0$ into the clause, which enhances the ability of the clause to discriminate between different patterns. For the other configurations, the feedback is always inaction (with probabilities $1.0$). In particular, if the clause evaluates to $0$, Type II Feedback provides no further stimuli for including literals of value $0$. Instead, reinforcing include further is left to the Type I Feedback. The purpose of this is to avoid local optima. In this way, the Type II Feedback supresses false positive output. }

Note that the formation of patterns is founded on frequent pattern mining. That is, the parameter $s$ above controls the granularity of the clauses. A larger $s$ allows more literals to be included in each clause, making the corresponding sub-patterns more fine-grained. 

{\color{black}A detailed analysis of the two feedback tables can be found in \cite{granmo2018tsetlin}.} We now introduce the training process in detail, focusing specifically on the one-bit case.

\begin{table}[h!]
\centering
\begin{tabular}{c|ccccc}
\multicolumn{2}{r|}{{\it Value of the clause} $C^i_j(X)$ }&\multicolumn{2}{c}{\True}&\multicolumn{2}{c}{\False}\\  
\multicolumn{2}{r|}{{\it Value of the Literal} $x_k$/$\lnot x_k$}&{\True}&{\False}&{\True}&{\False}\\
 \hline
 \hline
    \multirow{3}{*}{\bf  Include Literal}&\multicolumn{1}{c|}{$P(\mathrm{Reward})$}&$\frac{s-1}{s}$&NA&$0$&$0$\\
    &\multicolumn{1}{c|}{$P(\mathrm{Inaction})$}&$\frac{1}{s}$&NA&$\frac{s-1}{s}$&$\frac{s-1}{s}$\\
  &\multicolumn{1}{c|}{$P(\mathrm{Penalty})$}&$0$&NA&$\frac{1}{s} $&$\frac{1}{s}$\\
  \hline
  \multirow{3}{*}{\bf Exclude Literal }&\multicolumn{1}{c|}{$P(\mathrm{Reward})$}&$0$&$\frac{1}{s}$&$\frac{1}{s}$ &$\frac{1}{s}$\\
  &\multicolumn{1}{c|}{$P(\mathrm{Inaction})$}&$\frac{1}{s}$&$\frac{s-1}{s}$&$\frac{s-1}{s}$ &$\frac{s-1}{s}$\\
  &\multicolumn{1}{c|}{$P(\mathrm{Penalty})$}&$\frac{s-1}{s}$&$0$&$0$&$0$\\
  \hline
\end{tabular}
\caption{Type I Feedback --- Feedback upon receiving a sample with label $y=1$, for a single TA to decide whether to Include or Exclude a given literal $x_k/\neg x_k$ into $C^i_j$. NA means not applicable.}
\label{table:type_i_feedback}
\end{table}

\begin{table}[h!]
\centering
\begin{tabular}{c|ccccc}
\multicolumn{2}{r|}{\it Value of the clause $C^i_j(X)$}&\multicolumn{2}{c}{\True}&\multicolumn{2}{c}{\False}\\  
\multicolumn{2}{r|}{\it Value of the Literal $x_k/\neg x_k$}&{\True}&{\False}&{\True}&{\False}\\
 \hline
 \hline
    \multirow{3}{*}{\bf Include Literal }&\multicolumn{1}{c|}{$P(\mathrm{Reward})$}&$0$&$\mathrm{NA}$&$0$&$0$\\
    &\multicolumn{1}{c|}{$P(\mathrm{Inaction})$}&$1.0$&$\mathrm{NA}$&$1.0$&$1.0$\\
  &\multicolumn{1}{c|}{$P(\mathrm{Penalty})$}&$0$&$\mathrm{NA}$&$0$&$0$\\
  \hline
  \multirow{3}{*}{\bf Exclude Literal }&\multicolumn{1}{c|}{$P(\mathrm{Reward})$}&$0$&$0$&$0$&$0$\\
  &\multicolumn{1}{c|}{$P(\mathrm{Inaction})$}&$1.0$&$0$&$1.0$ &$1.0$\\
  &\multicolumn{1}{c|}{$P(\mathrm{Penalty})$}&$0$&$1.0$&$0$&$0$\\
  \hline
\end{tabular}
\caption{Type II Feedback --- Feedback upon receiving a sample with label $y=0$, for a single TA to decide whether to Include or Exclude a given literal $x_k/\neg x_k$ into $C^i_j$. NA means not applicable.}
\label{table:type_ii_feedback}
\end{table}

\subsubsection{The training process in the one-bit case}
\label{sec:onebittraning}
In the one-bit case, the input is reduced to a single variable $X=x$, which takes one of two values, $x \in \{0, 1\}$. Each example $x$ in the training set $\mathcal S$, in turn, has two types of labels, i.e., $y=0$ or $y=1$. This results in four input-label configurations: $(x, y) = (0,0)$, $(x, y) = (0,1)$, $(x, y) = (1,0)$, and $(x, y) = (1,1)$. Based on the probabilities that these input-label pairs appear in the training data set, we define the following two cases:
\\
\begin{itemize}
    \item {\bf Noise-free case:} \begin{itemize}
        \item $P(y=1|x=1)=1$, \item $P(y=0|x=1)=0$,
        \item $P(y=1|x=0)=0$, \item $P(y=0|x=0)=1$. 
    \end{itemize}
    \item {\bf Noisy case,} where $a,b \in (0,1)$:
    \begin{itemize}
        \item $P(y=1|x=1)=a$, \item $P(y=0|x=1)=1-a$;    
        \item $P(y=1|x=0)=b$, \item  $P(y=0|x=0)=1-b$. 
    \end{itemize}
\end{itemize}
For the sake of simplicity, and without loss of generality, we present the training process based on the noise-free case. In the noise-free case, the output $y=1$ is always associated with the input value $x=1$. That is, only one sub-pattern is associated with output $y=1$. We thus set the number of clauses $m=1$, meaning that the resulting TM only consists of one TA team. Accordingly, since we only have one input variable, $x$, the TA team contains two TAs. We thus have simplified the entire system into a game between two TAs.
    
We further define the two operators we are targeting in this paper, IDENTITY and NOT, as:
\begin{align}
    \mathrm{IDENTITY}(x) ~=&~ x, \nonumber \\
    \mathrm{NOT} (x) ~=&~ 1-x. \nonumber 
\end{align}
Accordingly, the above noise-free case covers the IDENTITY-operator, which can be converted into the NOT-operator by inverting the output $y$. 
We now describe the training process for the noise-free case step-by-step:

\begin{enumerate} 
\item We initialize the TAs by giving each of them a random state among the states associated with the exclude action.

\item \label{step:Clause} We obtain a new training sample $(x, y)$ and evaluate our single clause $C$ according to Eq. (\ref{eqn:clause1}). Depending on the actions of the two TAs, we get the following four possible clause configurations:
{\color{black}
{\footnotesize
\begin{subequations}
\label{eqn:C1all}
\begin{empheq} [left={C(x)=\empheqlbrace\,}]{align}
&I(x) \wedge I(\neg x) =0 \label{eqn:C1_1}\\
&I(x) \wedge E(\neg x)  =I(x)  = x \label{eqn:C1_2}\\
&E(x) \wedge I(\neg x)  =I(\neg x)  = \neg x \label{eqn:C1_3}\\
&E(x) \wedge E(\neg x)  = 1. \label{eqn:C1_4}
\end{empheq}
\end{subequations}
}
}

\item Based on the label $y$, the clause value $C(x)$, and the value of each individual literal $x$ or $\neg x$, we update the states of the TAs according to Table \ref{table:type_i_feedback} and Table \ref{table:type_ii_feedback}. The details on how the states of the TAs are updated will be elaborated in Section \ref{sect:noisefree}.

\item Repeat from Step \ref{step:Clause} until a given stopping criteria is met.

\end{enumerate}

Note that the training process defined here does not involve any ``voting'', this is because the targeted TM only involves a single clause. This simplification will make it easier for us to capture the core of the learning mechanism underlying a TM, i.e., how the feedback tables are able to guide the TAs to converge to the expected pattern.

\section{Analytical Procedure and Main Results for One-bit Case} \label{sect:one-bitcase}
In this section, we analyze the convergence of the TM in the one-bit case. The target of the analysis is to decide whether the clause $C$ converges to the correct relation, reflected by the training samples.  
This is equivalent to saying that we are to find out whether the clause converges to $C= x$ if and only if  $P(y = \True | x=1 ) > P(y= \False | x=1)$, or the clause converges to $C=\neg x$ if and only if  $P(y = \True | x=0 ) > P(y= \False | x=0)$.

\subsection{The noise-free case}
\label{sect:noisefree}
Recall that the samples are distributed according to the following probabilities for the noise-free case: $P(y=1|x=1)=1$ which implies $P(y=0|x=1)=0$, and $P(y=0|x=0)=1$ which implies $P(y=1|x=0)=0$. There are hence two types of samples, i.e., $(x,y)=(1,1)$ and $(x,y)=(0,0)$, and it is the IDENTITY relation that captures this pattern. We further assume that the input $x=1$ appears in the training data set with probability $P(x=1)=c$, which means that $x=0$ appears with probability $P(x=0)=1-c$. We are now ready to prove that the TM can converge to the IDENTITY relation after training.

\begin{mytheorem}
\label{Lemma:onebitAND}
In the noise-free case where $P(y=1|x=1)=1$ and $P(y=0|x=0)=1$, the TM converges surely to the IDENTITY relation in an infinite time horizon, i.e.,
\begin{align}
C(x) \rightarrow I(x) \wedge E(\neg x)=x,~ &{\text{when }}c\in(0,1].
\end{align} 
\end{mytheorem}

{\bf Proof:} The TM studied in this article consists of two TAs: $\mathrm{TA}_1$ for literal $x$ and $\mathrm{TA}_2$ for literal $\neg x$. There are thus four possible action configurations: $((E(x), E(\neg x))$, $(E(x), I(\neg x))$, $(I(x), E(\neg x))$, or $(I(x), I(\neg x))$. By combining these four configurations with the two types of training samples in the noise-free case, we get altogether eight scenarios. Table \ref{table:noise-free} displays
the state transitions of $\mathrm{TA}_1$ and $\mathrm{TA}_2$ for each of these scenarios\footnote{We have simplified all the state transition chains in the paper to a chain depth of $2$. However, our analysis addresses the general depth $N$.}. The state transitions are generated from the Type I and Type II Feedback tables, showing how $\mathrm{TA}_1$ and $\mathrm{TA}_2$ move from one state to another. We use Scenario 2 and Scenario 5 to exemplify how the state transitions are generated.

In Scenario 2, we assume that the current states of the TAs are $S_1, S_2 \in (N+1,2N)$. This means that both TAs select the exclude action, i.e.,  $\mathrm{TA}_1$  selects action $E(x)$ and $\mathrm{TA}_2$ selects action $E(\neg x)$. According to Eq. (\ref{eqn:C1_4}), the resulting clause becomes $C(x)=E(x)\wedge E(\neg x) =1$. When the sample $(0, 0)$ is fed to the system, we have $C(x)=1$. Label $y=0$ triggers Type II Feedback. We thus refer to Table \ref{table:type_ii_feedback} for the reward/inaction/penalty probabilities:
\begin{itemize}
\item We first consider $\mathrm{TA}_1$. As seen in the column where $C=1$ and $x=0$, $\mathrm{TA}_1$ receives a penalty with probability $1$ for choosing the action ``Exclude Literal''. This means that $\mathrm{TA}_1$ changes state to $S_1-1$. Accordingly, the state change weakens action ``Exclude Literal'' by moving towards the ``Include Literal''-side of the state space. As the input $x=0$ appears with probability $1-c$, the effective penalty probability is $1-c$ in the state transition diagram.

\item We now examine feedback for $\mathrm{TA}_2$. The column where $C=1$ and $\neg x=1$ shows that $\mathrm{TA}_2$ receives inaction feedback with probability $1$ for choosing the action ``Exclude Literal''. The effective inaction probability is thus $1-c$. For the sake of simplicity, all the self loops for inaction probabilities are omitted in all the state transition diagrams, which makes plotting the state transition for $\mathrm{TA}_2$ in this scenario trivial, as the state of $\mathrm{TA}_2$ simply remains unchanged.
\end{itemize}

In Scenario 5, we assume the current state of $\mathrm{TA}_1$ is $S_1 \in (1,N)$, which makes the TA select action $I(x)$. We further assume that $\mathrm{TA}_2$ is in State $S_2 \in (N+1,2N)$, producing action $E(\neg x)$. Then, when the sample $(1, 1)$ is fed to the system, we evaluate the clause according to Eq. (\ref{eqn:C1_2}). We thus get $C(x)=I(x)\wedge E(\neg x)  = I(x) \wedge 1 = x = 1$. Since label $y=1$ triggers Type I Feedback, we refer to Table \ref{table:type_i_feedback} for the reward/inaction/penalty probabilities:
\begin{itemize}
\item We first consider $\mathrm{TA}_1$. According to the column where $C=1$ and $x=1$, $\mathrm{TA}_1$ will receive a reward with probability $\frac{s-1}{s}$ for choosing the action ``Include Literal''. This means that $\mathrm{TA}_1$ will change state to $S_1-1$ with probability $\frac{s-1}{s}$, reinforcing the ``Include Literal'' action. Because $P(x=1)=c$, the effective reward probability is $\frac{(s-1)c}{s}$, as shown by the state transition diagram. Conversely, $\mathrm{TA}_1$ does not change state with probability $\frac{1}{s}\times c=\frac{c}{s}$, according to the inaction probability for ``Include Literal''. 
\item We now examine feedback for $\mathrm{TA}_2$.  The column where $C=1$ and $\neg x=0$ shows that $\mathrm{TA}_2$ will receive a reward with probability $\frac{1}{s}$ for choosing the action ``Exclude Literal''. This means that $\mathrm{TA}_2$ changes state to $S_2+1$ with probability $\frac{1}{s}$, reinforcing the ``Exclude Literal'' action. Again, by multiplying the probability $c$ of observing $x=1$, the effective reward probability becomes $\frac{c}{s}$. The state of $\mathrm{TA}_2$ will thus remain unchanged with probability $\frac{(s-1)c}{s}$ according to the inaction probability for ``Exclude Literal''. 
\end{itemize}

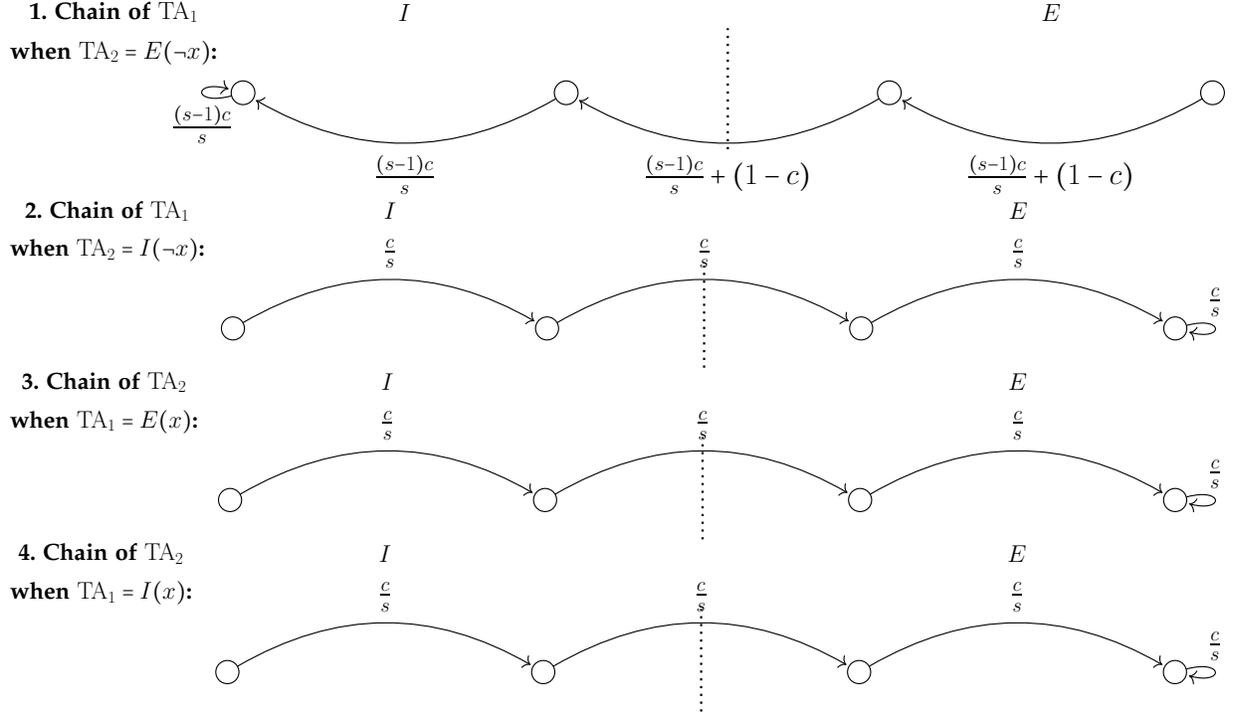
\begin{figure*}[htbp]
\begin{center}

\begin{minipage}{1\textwidth}
\resizebox{1\textwidth}{!}{
\begin{tikzpicture}[node distance = .35cm, font=\Huge]
    \tikzstyle{every node}=[scale=0.35]
    
    \node[state] (A) at (1,1) {};
    \node[state] (B) at (5,1) {};
    \node[state] (C) at (9,1) {};
    \node[state] (D) at (13,1) {};
    
    \node[thick] at (3,2) {$I$};
    \node[thick] at (11,2) {$E$};
    \node[thick] at (-0.6,2) {\bf 1. Chain of $\mathrm{TA}_1$};
    \node[thick] at (-0.6,1.5) {\bf when $\mathrm{TA}_2=E(\neg x)$:};
    
    \draw[dotted, thick] (7,0.3) -- (7,1.8);
    
    \draw[every loop]
    (A) edge[loop left] node [scale=1.2, below=0.1 of H] {$\frac{(s-1)c}{s}$} (A)
    (B) edge[bend left] node [scale=1.2, below=0.1 of H] {$\frac{(s-1)c}{s}$} (A) 
    (C) edge[bend left] node [scale=1.2, below=0.1 of H] {$\frac{(s-1)c}{s}+(1-c)$} (B) 
    (D) edge[bend left] node [scale=1.2, below=0.1 of C] {$\frac{(s-1)c}{s}+(1-c)$} (C);
    
\end{tikzpicture}
}

\resizebox{1\textwidth}{!}{
\begin{tikzpicture}[node distance = .35cm, font=\Huge]
    \tikzstyle{every node}=[scale=0.35]
    
    \node[state] (A) at (1,1) {};
    \node[state] (B) at (5,1) {};
    \node[state] (C) at (9,1) {};
    \node[state] (D) at (13,1) {};
    
    \node[thick] at (3,2.5) {$I$};
    \node[thick] at (11,2.5) {$E$};
    \node[thick] at (-0.6,2.5) {\bf 2. Chain of $\mathrm{TA}_1$};
    \node[thick] at (-0.6,2) {\bf when $\mathrm{TA}_2=I(\neg x)$:};
    
    \draw[dotted, thick] (7,0.5) -- (7,1.8);

     \draw[every loop]
    (A) edge[bend left] node [scale=1.2, above=0.1 of H] {$\frac{c}{s}$} (B)
    (B) edge[bend left] node [scale=1.2, above=0.1 of H] {$\frac{c}{s}$} (C) 
    (C) edge[bend left] node [scale=1.2, above=0.1 of H] {$\frac{c}{s}$} (D) 
    (D) edge[loop right] node [scale=1.2, above=0.1 of C] {$\frac{c}{s}$} (D);

\end{tikzpicture}
}

\resizebox{1\textwidth}{!}{
\begin{tikzpicture}[node distance = .35cm, font=\Huge]
    \tikzstyle{every node}=[scale=0.35]
    
    \node[state] (A) at (1,1) {};
    \node[state] (B) at (5,1) {};
    \node[state] (C) at (9,1) {};
    \node[state] (D) at (13,1) {};
    
    \node[thick] at (3,2.5) {$I$};
    \node[thick] at (11,2.5) {$E$};
    \node[thick] at (-0.6,2.5) {\bf 3. Chain of $\mathrm{TA}_2$};
    \node[thick] at (-0.6,2) {\bf when $\mathrm{TA}_1=E(x)$:};
    
    \draw[dotted, thick] (7,0.5) -- (7,1.8);

     \draw[every loop]
    (A) edge[bend left] node [scale=1.2, above=0.1 of H] {$\frac{c}{s}$} (B)
    (B) edge[bend left] node [scale=1.2, above=0.1 of H] {$\frac{c}{s}$} (C) 
    (C) edge[bend left] node [scale=1.2, above=0.1 of H] {$\frac{c}{s}$} (D) 
    (D) edge[loop right] node [scale=1.2, above=0.1 of C] {$\frac{c}{s}$} (D);

\end{tikzpicture}
}

\resizebox{1\textwidth}{!}{
\begin{tikzpicture}[node distance = .35cm, font=\Huge]
    \tikzstyle{every node}=[scale=0.35]
    
    \node[state] (A) at (1,1) {};
    \node[state] (B) at (5,1) {};
    \node[state] (C) at (9,1) {};
    \node[state] (D) at (13,1) {};
    
    \node[thick] at (3,2.5) {$I$};
    \node[thick] at (11,2.5) {$E$};
    \node[thick] at (-0.6,2.5) {\bf 4. Chain of $\mathrm{TA}_2$};
    \node[thick] at (-0.6,2) {\bf when $\mathrm{TA}_1=I(x)$:};
    
    \draw[dotted, thick] (7,0.5) -- (7,1.8);
    
     \draw[every loop]
    (A) edge[bend left] node [scale=1.2, above=0.1 of H] {$\frac{c}{s}$} (B)
    (B) edge[bend left] node [scale=1.2, above=0.1 of H] {$\frac{c}{s}$} (C) 
    (C) edge[bend left] node [scale=1.2, above=0.1 of H] {$\frac{c}{s}$} (D) 
    (D) edge[loop right] node [scale=1.2, above=0.1 of C] {$\frac{c}{s}$} (D);

\end{tikzpicture}
}
\end{minipage}

\end{center}
\caption{\label{fig:noisefreecompact} Markov chains in the noise-free case, when $ P(y=1 | X=1) = 1$, $P(y=0 | X=0) = 1$, and $P(X=1)=c$, $c \in (0,1)$.}
\end{figure*}

For the other scenarios, the state transitions for $\mathrm{TA}_1$ and $\mathrm{TA}_2$ can be generated in the same manner. Furthermore, based on Table \ref{table:noise-free}, the state transitions of $\mathrm{TA}_1$ and $\mathrm{TA}_2$ can be combined to Markov chains, as shown in Figure \ref{fig:noisefreecompact}. We analyze the convergence of the TM by investigating the Markov chains.

\begin{itemize}
\item We first consider $0<c\leq 1$, i.e., the case with  two types of samples $(1,1)$ and $(0,0)$, or only one type, $(1,1)$. The analysis can then be conducted in two steps:
\begin{enumerate}
    \item As $s>1$ by definition, we have $\frac{c}{s}>0$. The Markov chains for $\mathrm{TA}_2$ shows that no matter which action $\mathrm{TA}_1$ takes, $\mathrm{TA}_2$ moves and converges to $E(\neg x)$.
    \item When $\mathrm{TA}_2$ converges to $E(\neg x)$, Chain 1 shows that $\mathrm{TA}_1$ converges to $I(x)$.
\end{enumerate}
We thus can confirm that when $0<c \leq 1$, $C(x) \rightarrow I(x)\wedge E(\neg x) =x$, reflecting the IDENTITY relation.

\item The remaining case is $c=0$, i.e., when there is only one type of samples $(0,0)$. Again, the analysis consists of two steps:
\begin{enumerate}
    \item The Markov chain for $\mathrm{TA}_2$ shows that its state does not change, meaning the final state of $\mathrm{TA}_2$ depends only on its initial state. $\mathrm{TA}_2$ can thus be choosing either $I(\neg x)$ or $E(\neg x)$.
    
    \item If $\mathrm{TA}_2=E(\neg x)$, Chain 1 shows that $\mathrm{TA}_1$ converges to $I(x)$ with probability $1$, and we get $C\rightarrow I(x)\wedge E(\neg x)=x$. However, if $\mathrm{TA}_2=I(\neg x)$, then Chain 2 shows that the decision of $\mathrm{TA}_1$ depends fully on its initial state, which can be either $I(x)$ or $E(x)$.
\end{enumerate}
In other words, when $c=0$, $C(x)$ does not converge consistently to $I(x)\wedge E(\neg x) =x$. This is simply because there are no samples where $y=1$ to learn from. 
\end{itemize}
In summary, as long as there are samples with label $y=1$, $C(x)$ will converge to $I(x) \wedge E(\neg x)=x$, reflecting the IDENTITY relation in pattern $y=1$. Theorem \ref{Lemma:onebitAND} is thus proven. \hfill$\blacksquare$ 

\begin{remark}
\label{remark:onebit}
Theorem \ref{Lemma:onebitAND} does not have any requirements on the depths of the TAs. The TAs can learn the correct pattern even if $N=1$.
\end{remark}

\subsection{The noisy case}
In the case with noise, we assume the samples are distributed according to the probabilities: $P(y=1|x=1)=a$, which implies $P(y=0|x=1)=1-a$, and $P(y=1|x=0)=b$, which implies $P(y=0|x=0)=1-b$.  There are hence four types of samples, i.e., $(1,1)$, $(1,0)$, and $(0,0)$, $(1,0)$. These provide two types of candidate relations for predicting $y$, either IDENTITY or NOT.  
We are to find out whether the TM can converge to the correct relation over an infinite training horizon.

\begin{figure*}
\centering
\begin{minipage}{1\textwidth}
\resizebox{1\textwidth}{!}{
\begin{tikzpicture}[node distance = .35cm, font=\Huge]
    \tikzstyle{every node}=[scale=0.35]
    
    \node[state] (A) at (0,2) {~~~0~~~~};
    \node[state] (B) at (1.5,2) {~~~1~~~~};
    
    \node[state,draw=white] (M) at (3,2) {~~~$....$~~~};
    
    \node[state] (C) at (4.5,2) {$N-1$};
    \node[state] (D) at (6,2) {~~~\!$N$~~~};
    
    \node[state] (E) at (7.5,2) {$N+1$};
    \node[state] (F) at (9,2) {$N+2$};
    
    \node[state,draw=white] (G) at (10.5,2) {~~~$....$~~~};
    
    \node[state] (H) at (12,2) {\!$2N-1$};
    \node[state] (I) at (13.5,2) {~~\!$2N$~~~};
    
    \node[thick] at (0,0.7) {$\pi_{0}$};
    \node[thick] at (1.5,0.7) {$\pi_{1}$};
    
    \node[thick] at (3,0.7) {~~~~~$....$~~~~~~};
	
    \node[thick] at (4.5,0.7) {$\pi_{N-1}$};
    \node[thick] at (6,0.7) {$\pi_{N}$};
    
    \node[thick] at (7.5,0.7) {$\pi_{N+1}$};
    \node[thick] at (9,0.7) {$\pi_{N+2}$};
    
    \node[thick] at (10.5,0.7) {~~~~~$....$~~~~~~};
    
    \node[thick] at (12,0.7) {$\pi_{2N-1}$};
    \node[thick] at (13.5,0.7) {$\pi_{2N}$};
    
    \node[thick] at (4,3.5) {Include};
    \node[thick] at (9.5,3.5) {Exclude};

    \draw[every loop]
    (A) edge[bend left] node [scale=1.2, above=0.1 of B]{$\alpha$} (B)
    (B) edge[bend left] node  [scale=1.2, above=0.1 of M] {$\alpha$} (M)
    (M) edge[bend left] node  [scale=1.2, above=0.1 of C] {$\alpha$} (C)
    (C) edge[bend left] node [scale=1.2, above=0.1 of D] {$\alpha$} (D)
    (D) edge[bend left] node  [scale=1.2, above=0.1 of E] {$\alpha$} (E)
    (E) edge[bend left] node  [scale=1.2, above=0.1 of F] {$\alpha$} (F)
    (F) edge[bend left] node [scale=1.2, above=0.1 of G] {$\alpha$} (G)
    (G) edge[bend left] node [scale=1.2, above=0.1 of H] {$\alpha$} (H)
    (H) edge[bend left] node [scale=1.2, above=0.1 of I] {$\alpha$} (I);
    
    \draw[every loop]
    (I) edge[bend left] node [scale=1.2, below=0.1 of H] {$\gamma$} (H)
    (H) edge[bend left] node  [scale=1.2, below=0.1 of G] {$\gamma$} (G)
    (G) edge[bend left] node [scale=1.2, below=0.1 of F] {$\gamma$} (F)
    (F) edge[bend left] node  [scale=1.2, below=0.1 of E] {$\gamma$} (E)
    (E) edge[bend left] node  [scale=1.2, below=0.1 of D] {$\gamma$} (D)
    (D) edge[bend left ] node [scale=1.2, below=0.1 of C] {$\beta$} (C)
    (C) edge[bend left] node  [scale=1.2, below=0.1 of M] {$\beta$} (M)
    (M) edge[bend left] node  [scale=1.2, below=0.1 of C] {$\beta$} (B)
    (B) edge[bend left ] node [scale=1.2, below=0.1 of A] {$\beta$} (A);
    
      \draw[dotted, thick] (6.75,1) -- (6.75,3);
\end{tikzpicture}
}
\end{minipage}
\caption{A Markov chain for Lemma \ref{Lemma:staticanalysis}. }
\label{MC2}
\end{figure*}
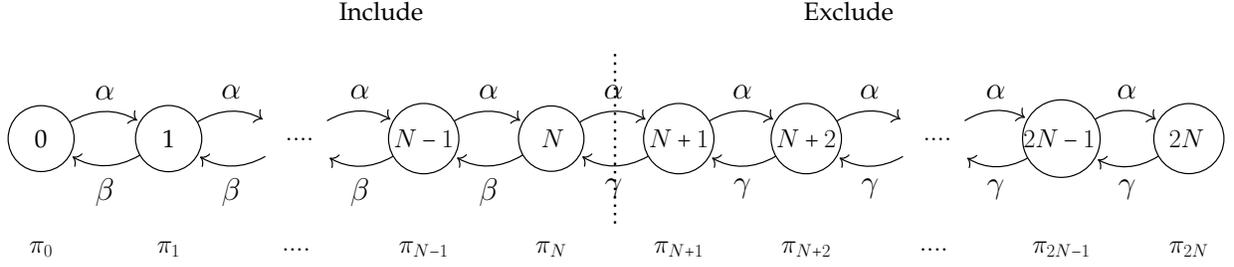

We start with the static state analysis of the Markov chain shown in Figure \ref{MC2}. The Markov chain is indeed a two-action TA that has $2N$-state, with $\alpha,~ \beta$, and $\gamma$ being the transition probabilities between states. Note that we have left out the self-loops of the Markov chain in the diagram. We are to derive the conditions for the TA to converge to one of the actions with probability $1$. 

\begin{mylemma}
\label{Lemma:staticanalysis}
When $N\rightarrow \infty$, action ``Include'' will be selected with probability 1 if $\alpha < \min(\beta, \gamma)$. Similarly, action ``Exclude'' will be selected with probability 1 if $\alpha>\max(\beta, \gamma)$.
\end{mylemma}

{\bf Proof:} 
Denoting the static state probability of State $i$ as $\pi_i$, $i=0, 1,..., 2N$, the balance equations of the Markov chain are:
\begin{align}
\alpha \pi_0=\beta \pi_1&\implies\pi_1=\frac{\alpha}{\beta} \pi_0, \nonumber\\
\alpha\pi_1=\beta\pi_2&\implies\pi_2=\frac{\alpha}{\beta}\pi_1=\left(\frac{\alpha}{\beta}\right)^2\pi_0,
\nonumber\\
&\ldots,
\nonumber\\
\alpha\pi_{N-1}=\beta\pi_N&\implies\pi_N=\frac{\alpha}{\beta}\pi_{N-1}=\left(\frac{\alpha}{\beta}\right)^N\pi_0, 
\nonumber\\
\alpha\pi_N=\gamma\pi_{N+1}&\implies\pi_{N+1}=\frac{\alpha}{\gamma}\pi_N=\frac{\alpha}{\gamma}\left(\frac{\alpha}{\beta}\right)^N\pi_0,
\nonumber\\
\alpha\pi_{N+1}=\gamma\pi_{N+2}&\implies\pi_{N+2}=\frac{\alpha}{\gamma}\pi_{N+1}=\left(\frac{\alpha}{\gamma}\right)^2\left(\frac{\alpha}{\beta}\right)^N\pi_0,
\nonumber\\
&\ldots,
\nonumber\\
\alpha\pi_{2N-1}=\gamma\pi_{2N}&\implies\pi_{2N}=\frac{\alpha}{\gamma}\pi_{2N-1}=\left(\frac{\alpha}{\gamma}\right)^N\left(\frac{\alpha}{\beta}\right)^N\pi_0.
\label{BDP2}
\end{align}
As $(\pi_0+\pi_1+\pi_2+\ldots+\pi_{N}+\pi_{N+1}+\ldots+\pi_{2N})=1$, we have:
{\footnotesize
\begin{align}
&\pi_0\left(1+\frac{\alpha}{\beta}+\left(\frac{\alpha}{\beta}\right)^2+\ldots+\left(\frac{\alpha}{\beta}\right)^N\right)+\nonumber\\
&\pi_0\left(\frac{\alpha}{\beta}\right)^N\left(\left(\frac{\alpha}{\gamma}\right)+\left(\frac{\alpha}{\gamma}\right)^2+\ldots+\left(\frac{\alpha}{\gamma}\right)^{N}\right)=1\nonumber\\
\implies&\pi_0\left(\frac{1-\left(\frac{\alpha}{\beta}\right)^{(N+1)}}{1-\frac{\alpha}{\beta}}+\left(\frac{\alpha}{\beta}\right)^N\left(\frac{1-\left(\frac{\alpha}{\gamma}\right)^{N+1}}{1-\frac{\alpha}{\gamma}} \right)  \right)=1.
\label{eq:BDP3}
\end{align}
}
If $\alpha< \min(\beta, \gamma)$, then when $N\rightarrow \infty$, Eq. (\ref{eq:BDP3}) is 
$\pi_0 \left(\frac{1}{1-\frac{\alpha}{\beta}}\right) = 1$, which implies $\pi_0=1-\frac{\alpha}{\beta}$. The probability that the state of the Markov chain stays on the left-hand side, can thus be calculated as $\lim\limits_{N\rightarrow \infty}(\pi_0+\pi_1+\ldots+\pi_N)=\lim\limits_{N\rightarrow \infty}\pi_0\left(1+\frac{\alpha}{\beta}+\left(\frac{\alpha}{\beta}\right)^2+\ldots+\left(\frac{\alpha}{\beta}\right)^N\right)=\lim\limits_{N\rightarrow \infty}\pi_0\frac{1-\left(\frac{\alpha}{\beta}\right)^{(N+1)}}{1-\frac{\alpha}{\beta}}=\pi_0\left(\frac{1}{1-\frac{\alpha}{\beta}}\right)=1$. Accordingly, we have proved that action ``Include'' will be selected with probability $1$ if $\alpha < \min\{\beta, \gamma\}$ when $N\rightarrow \infty$.

The above procedure can be modified to prove that action ``Exclude'' will be selected with probability $1$ if $\alpha>\max\{\beta, \gamma\}$ when $N\rightarrow \infty$.
\hfill$\blacksquare$ 

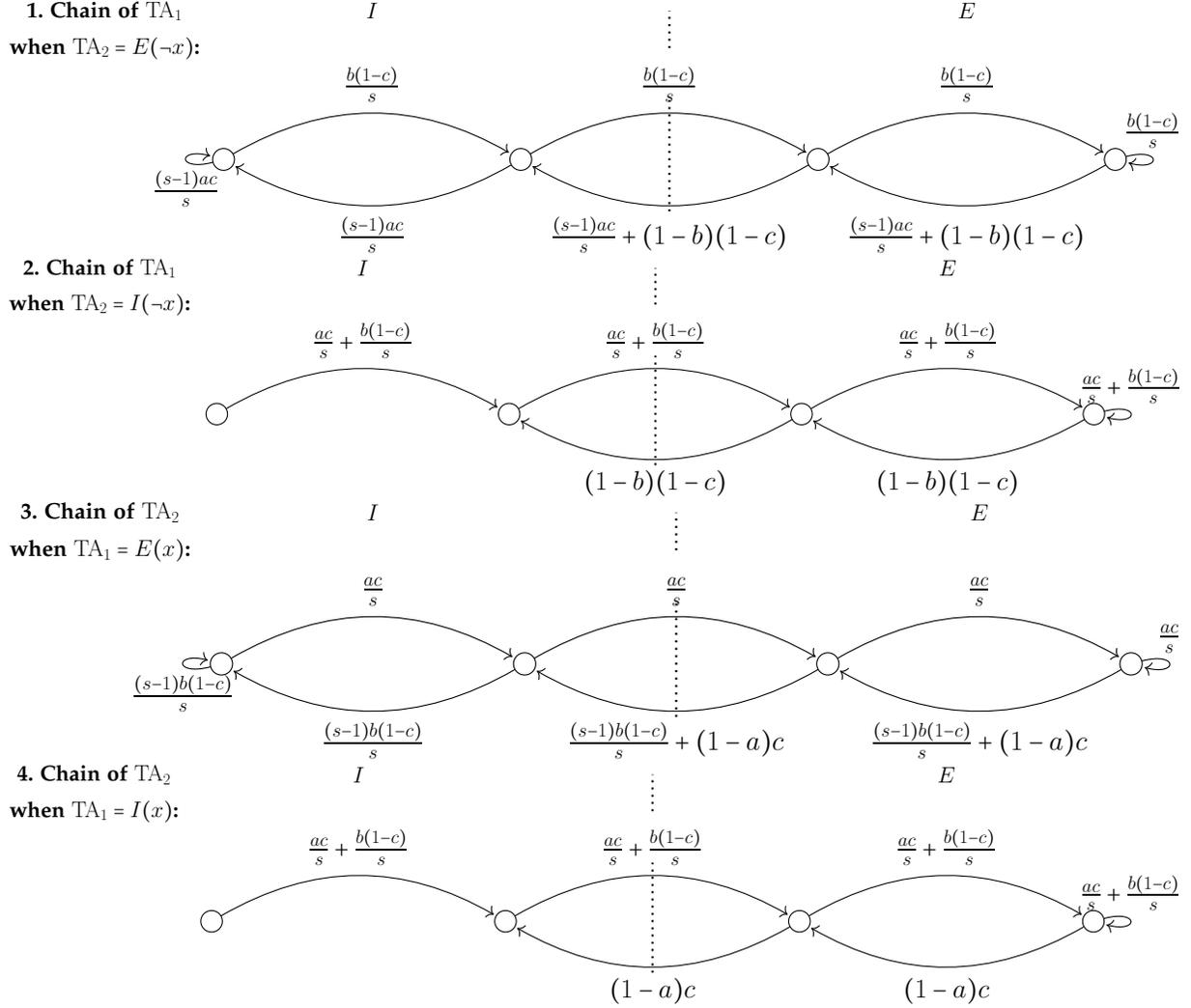
\begin{figure*}[htbp]
\begin{center}
\begin{minipage}{1\textwidth}
\resizebox{1\textwidth}{!}{
\begin{tikzpicture}[node distance = .35cm, font=\Huge]
    \tikzstyle{every node}=[scale=0.35]
    
    \node[state] (A) at (1,1) {};
    \node[state] (B) at (5,1) {};
    \node[state] (C) at (9,1) {};
    \node[state] (D) at (13,1) {};
    
    \node[thick] at (3,3) {$I$};
    \node[thick] at (11,3) {$E$};
    \node[thick] at (-0.6,3) {\bf 1. Chain of $\mathrm{TA}_1$};
    \node[thick] at (-0.6,2.5) {\bf when $\mathrm{TA}_2=E(\neg x)$:};
    
    \draw[dotted, thick] (7,0.3) -- (7,1.8);
    \draw[dotted, thick] (7,2.5) -- (7,3);
    
    \draw[every loop]
    (A) edge[loop left] node [scale=1.2, below=0.1 of H] {$\frac{(s-1)ac}{s}$} (A)
    (B) edge[bend left] node [scale=1.2, below=0.1 of H] {$\frac{(s-1)ac}{s}$} (A) 
    (C) edge[bend left] node [scale=1.2, below=0.1 of H] {$\frac{(s-1)ac}{s}+(1-b)(1-c)$} (B) 
    (D) edge[bend left] node [scale=1.2, below=0.1 of C] {$\frac{(s-1)ac}{s}+(1-b)(1-c)$} (C);
    
     \draw[every loop]
    (A) edge[bend left] node [scale=1.2, above=0.1 of H] {$\frac{b(1-c)}{s}$} (B)
    (B) edge[bend left] node [scale=1.2, above=0.1 of H] {$\frac{b(1-c)}{s}$} (C) 
    (C) edge[bend left] node [scale=1.2, above=0.1 of H] {$\frac{b(1-c)}{s}$} (D) 
    (D) edge[loop right] node [scale=1.2, above=0.1 of C] {$\frac{b(1-c)}{s}$} (D);

\end{tikzpicture}
}

\resizebox{1\textwidth}{!}{
\begin{tikzpicture}[node distance = .35cm, font=\Huge]
    \tikzstyle{every node}=[scale=0.35]
    
    \node[state] (A) at (1,1) {};
    \node[state] (B) at (5,1) {};
    \node[state] (C) at (9,1) {};
    \node[state] (D) at (13,1) {};
    
    \node[thick] at (3,3) {$I$};
    \node[thick] at (11,3) {$E$};
    \node[thick] at (-0.6,3) {\bf 2. Chain of $\mathrm{TA}_1$};
    \node[thick] at (-0.6,2.5) {\bf when $\mathrm{TA}_2=I(\neg x)$:};
    
    \draw[dotted, thick] (7,0.3) -- (7,1.8);
    \draw[dotted, thick] (7,2.5) -- (7,3);
    
    \draw[every loop]
    (C) edge[bend left] node [scale=1.2, below=0.1 of H] {$(1-b)(1-c)$} (B) 
    (D) edge[bend left] node [scale=1.2, below=0.1 of C] {$(1-b)(1-c)$} (C);
    
     \draw[every loop]
    (A) edge[bend left] node [scale=1.2, above=0.1 of H] {$\frac{ac}{s}+\frac{b(1-c)}{s}$} (B)
    (B) edge[bend left] node [scale=1.2, above=0.1 of H] {$\frac{ac}{s}+\frac{b(1-c)}{s}$} (C) 
    (C) edge[bend left] node [scale=1.2, above=0.1 of H] {$\frac{ac}{s}+\frac{b(1-c)}{s}$} (D) 
    (D) edge[loop right] node [scale=1.2, above=0.1 of C] {$\frac{ac}{s}+\frac{b(1-c)}{s}$} (D);

\end{tikzpicture}
}

\resizebox{1\textwidth}{!}{
\begin{tikzpicture}[node distance = .35cm, font=\Huge]
    \tikzstyle{every node}=[scale=0.35]
    
    \node[state] (A) at (1,1) {};
    \node[state] (B) at (5,1) {};
    \node[state] (C) at (9,1) {};
    \node[state] (D) at (13,1) {};
    
    \node[thick] at (3,3) {$I$};
    \node[thick] at (11,3) {$E$};
    \node[thick] at (-0.6,3) {\bf 3. Chain of $\mathrm{TA}_2$};
    \node[thick] at (-0.6,2.5) {\bf when $\mathrm{TA}_1=E(x)$:};
    
    \draw[dotted, thick] (7,0.3) -- (7,1.8);
    \draw[dotted, thick] (7,2.5) -- (7,3);
    
    \draw[every loop]
    (A) edge[loop left] node [scale=1.2, below=0.1 of H] {$\frac{(s-1)b(1-c)}{s}$} (A)
    (B) edge[bend left] node [scale=1.2, below=0.1 of H] {$\frac{(s-1)b(1-c)}{s}$} (A) 
    (C) edge[bend left] node [scale=1.2, below=0.1 of H] {$\frac{(s-1)b(1-c)}{s}+(1-a)c$} (B) 
    (D) edge[bend left] node [scale=1.2, below=0.1 of C] {$\frac{(s-1)b(1-c)}{s}+(1-a)c$} (C);
    
     \draw[every loop]
    (A) edge[bend left] node [scale=1.2, above=0.1 of H] {$\frac{ac}{s}$} (B)
    (B) edge[bend left] node [scale=1.2, above=0.1 of H] {$\frac{ac}{s}$} (C) 
    (C) edge[bend left] node [scale=1.2, above=0.1 of H] {$\frac{ac}{s}$} (D) 
    (D) edge[loop right] node [scale=1.2, above=0.1 of C] {$\frac{ac}{s}$} (D);

\end{tikzpicture}
}

\resizebox{1\textwidth}{!}{
\begin{tikzpicture}[node distance = .35cm, font=\Huge]
    \tikzstyle{every node}=[scale=0.35]
    
    \node[state] (A) at (1,1) {};
    \node[state] (B) at (5,1) {};
    \node[state] (C) at (9,1) {};
    \node[state] (D) at (13,1) {};
    
    \node[thick] at (3,3) {$I$};
    \node[thick] at (11,3) {$E$};
    \node[thick] at (-0.6,3) {\bf 4. Chain of $\mathrm{TA}_2$};
    \node[thick] at (-0.6,2.5) {\bf when $\mathrm{TA}_1=I(x)$:};
    
    \draw[dotted, thick] (7,0.3) -- (7,1.8);
    \draw[dotted, thick] (7,2.5) -- (7,3);
    
    \draw[every loop]
    (C) edge[bend left] node [scale=1.2, below=0.1 of H] {$(1-a)c$} (B) 
    (D) edge[bend left] node [scale=1.2, below=0.1 of C] {$(1-a)c$} (C);
    
     \draw[every loop]
    (A) edge[bend left] node [scale=1.2, above=0.1 of H] {$\frac{ac}{s}+\frac{b(1-c)}{s}$} (B)
    (B) edge[bend left] node [scale=1.2, above=0.1 of H] {$\frac{ac}{s}+\frac{b(1-c)}{s}$} (C) 
    (C) edge[bend left] node [scale=1.2, above=0.1 of H] {$\frac{ac}{s}+\frac{b(1-c)}{s}$} (D) 
    (D) edge[loop right] node [scale=1.2, above=0.1 of C] {$\frac{ac}{s}+\frac{b(1-c)}{s}$} (D);

\end{tikzpicture}
}
\end{minipage}

\end{center}
\caption{\label{fig:noisycasecompact} Markov chains in the noisy case, where $ P(y=1 | X=1) = a$, and $P(y=1 | X=0) = b$; $P(X=1)=c$; $a,b,c \in (0,1)$.}
\end{figure*}

We again use the Markov chain to analyze the convergence of the TM. Following the same process as in the noise-free case, we plot the state transitions for both TAs, per unique scenario. In the noisy case, there are in total $16$ scenarios. Table \ref{Table:noisycasepart1} and Table \ref{Table:noisycasepart2} list the state transitions for $\mathrm{TA}_1$ and $\mathrm{TA}_2$ in each of these scenarios. By combining the listed transitions, we get the Markov chains for $\mathrm{TA}_1$ and $\mathrm{TA}_2$, as shown in Figure \ref{fig:noisycasecompact}.

We examine $\mathrm{TA}_1$ first: 
\begin{enumerate}
\item When $\mathrm{TA}_2=E(\neg x)$, the state transition of $\mathrm{TA}_1$ is shown by the first Markov chain in Figure \ref{fig:noisycasecompact}. 
\begin{enumerate}
    \item In order for $\mathrm{TA}_1 \rightarrow E(x)$, by Lemma~\ref{Lemma:staticanalysis}, we must have:
{\footnotesize
    \begin{align}
    &\frac{b(1-c)}{s} > \nonumber\\
&\max \left(\frac{(s-1)ac}{s},~ \frac{(s-1)ac}{s}+(1-b)(1-c)\right)\nonumber\\
    &\implies
    \frac{b(1-c)}{s} > \frac{(s-1)ac}{s}+(1-b)(1-c)\nonumber\\
    &\implies
     s<\frac{ac+b(1-c)}{ac+(1-b)(1-c)} \xrightarrow{\text{define to be}}s_1
    \end{align}
}
    
    \item In order for $\mathrm{TA}_1 \rightarrow I(x)$, by Lemma \ref{Lemma:staticanalysis}, we need to have: 
{\footnotesize
    \begin{align}
    &\frac{b(1-c)}{s} < \nonumber\\
&\min \left(\frac{(s-1)ac}{s}, ~ \frac{(s-1)ac}{s}+(1-b)(1-c)\right) \nonumber\\
    &\implies 
     \frac{b(1-c)}{s} < \frac{(s-1)ac}{s} \nonumber\\
    &\implies
     s>\frac{ac+b(1-c)}{ac}\xrightarrow{\text{define to be}}s_2
    \end{align}
}
\end{enumerate}

\item When $\mathrm{TA}_2=I(\neg x)$, the state transition of $\mathrm{TA}_1$ is shown by the second Markov chain in Figure \ref{fig:noisycasecompact}.
\begin{enumerate}
    \item In order for $\mathrm{TA}_1 \rightarrow E(x)$, by Lemma~ \ref{Lemma:staticanalysis}, we have: 
{\footnotesize
    \begin{align}
    &\frac{ac}{s}+\frac{b(1-c)}{s} > (1-b)(1-c) \nonumber\\
    \implies 
    & s<\frac{ac+b(1-c)}{(1-b)(1-c)}\xrightarrow{\text{define to be}}s_3
    \end{align}
}    
    \item $\mathrm{TA}_1$ will never converge to $I(x)$ when $\mathrm{TA}_2=I(\neg x)$.
\end{enumerate}
\end{enumerate}

We now apply the same analysis on $\mathrm{TA}_2$:
\begin{enumerate}
\item When $\mathrm{TA}_1=E(x)$, the state transition of $\mathrm{TA}_2$ is shown by the third Markov chain in Figure \ref{fig:noisycasecompact}. 
\begin{enumerate}
    \item According to Lemma \ref{Lemma:staticanalysis}, in order for $\mathrm{TA}_2 \rightarrow E(\neg x)$, we must have:
{\footnotesize
    \begin{align}
    &\frac{ac}{s} > \nonumber\\
&\max \left(\frac{(s-1)b(1-c)}{s}, \frac{(s-1)b(1-c)}{s}+(1-a)c\right)\nonumber\\
    &\implies
    \frac{ac}{s} > \frac{(s-1)b(1-c)}{s}+(1-a)c\nonumber\\
    &\implies
    s<\frac{ac+b(1-c)}{b(1-c)+(1-a)c}\xrightarrow{\text{define to be}}s_4
    \end{align}
}
    
    \item Similarly, for $\mathrm{TA}_2 \rightarrow I(\neg x)$, we have:
{\footnotesize
    \begin{align}
    &\frac{ac}{s} <  \nonumber\\
&\min\left(\frac{(s-1)b(1-c)}{s}, \frac{(s-1)b(1-c)}{s}+(1-a)c\right) \nonumber\\
    &\implies 
     \frac{ac}{s} < \frac{(s-1)b(1-c)}{s} \nonumber\\
    &\implies
     s>\frac{ac+b(1-c)}{b(1-c)}\xrightarrow{\text{define to be}}s_5
    \end{align}
}
\end{enumerate}

\item When $\mathrm{TA}_1=I(x)$, the state transition of $\mathrm{TA}_2$ is shown by the last Markov chain in Figure~\ref{fig:noisycasecompact}.
\begin{enumerate}
    \item For $\mathrm{TA}_2 \rightarrow E(\neg x)$, by Lemma \ref{Lemma:staticanalysis}, we have: 
{\footnotesize
    \begin{align}
    &\frac{ac}{s}+\frac{b(1-c)}{s} > (1-a)c \nonumber\\
    \implies 
    & s<\frac{ac+b(1-c)}{(1-a)c}\xrightarrow{\text{define to be}}s_6
    \end{align}
}
    
    \item $\mathrm{TA}_2$ will never converge to $I(\neg x)$ when $\mathrm{TA}_1=I(x)$.
\end{enumerate}

\end{enumerate}

We list here the definition of $s_1, s_2, s_3, s_4, s_5$ and $s_6$ for easy reference:
{\small
\begin{align}
&s_1=\frac{ac+b(1-c)}{ac+(1-b)(1-c)} \xrightarrow{{\text when}~c=0.5} s_1 =\frac{a+b}{a+1-b}, \nonumber\\
&s_2=\frac{ac+b(1-c)}{ac} ~~~~~~~\xrightarrow{{\text when}~c=0.5} s_2 =\frac{a+b}{a},\nonumber\\
&s_3=\frac{ac+b(1-c)}{(1-b)(1-c)} ~~~~~~\xrightarrow{{\text when}~c=0.5} s_3 =\frac{a+b}{1-b},\nonumber\\
&s_4=\frac{ac+b(1-c)}{b(1-c)+(1-a)c} \xrightarrow{{\text when}~c=0.5} s_4 =\frac{a+b}{b+1-a},\nonumber\\
&s_5=\frac{ac+b(1-c)}{b(1-c)} ~~~~~~~\xrightarrow{{\text when}~c=0.5} s_5 =\frac{a+b}{b},\nonumber\\
&s_6=\frac{ac+b(1-c)}{(1-a)c} ~~~~~~~\xrightarrow{{\text when}~c=0.5} s_6 =\frac{a+b}{1-a}.\nonumber
\end{align}
}
The conditions for $\mathrm{TA}_1$ and $\mathrm{TA}_2$ to converge to different actions are summarized in Table \ref{Table:S}, from which, we have:
\begin{enumerate}
\item For $(\mathrm{TA}_1,\mathrm{TA}_2)\rightarrow (E(x),E(\neg x))$, we need $s<\min(s_1,s_4)$. As $s\geq 1$ by definition, we have $\min(s_1,s_4)>1$ that implies $a>0.5,~b>0.5$. 
\item For $(\mathrm{TA}_1,\mathrm{TA}_2)\rightarrow (I(x),E(\neg x))$, we need $s_2<s<s_6$. In this case, $s_6$ must be greater than $s_2$, which implies $a>0.5$.
\item For $(\mathrm{TA}_1,\mathrm{TA}_2)\rightarrow (E(x),I(\neg x))$, we need $s_5<s<s_3$, which implies $s_5<s_3$ that leads to $b>0.5$.
\end{enumerate}

\begin{table}
\centering
\caption{Conditions of $s$ for TAs to converge.}
\label{Table:S}
\begin{tabular}[]{|l|l|l|}
\hline

\textbf{Actions of $\mathrm{TA}_2$} 
& \textbf{For $\mathrm{TA}_1 \rightarrow E(x)$:} 
& \textbf{For $\mathrm{TA}_1 \rightarrow I(x)$:} 
\\
\hline

$\mathrm{TA}_2=E(\neg x)$
&
$s<s_1$
&
$s>s_2$
\\
\hline

$\mathrm{TA}_2=I(\neg x)$
&
$s<s_3$
&
$s>s_3$ NA
\\
\hline
\hline

\textbf{Actions of $\mathrm{TA}_1$} 
& \textbf{For $\mathrm{TA}_2 \rightarrow E(\neg x)$:} 
& \textbf{For $\mathrm{TA}_2 \rightarrow I(\neg x)$:} 
\\
\hline

$\mathrm{TA}_1=E(x)$
&
$s<s_4$
&
$s>s_5$
\\
\hline

$\mathrm{TA}_1=I(x)$
&
$s<s_6$
&
$s>s_6$ NA
\\
\hline

\end{tabular}
\end{table}

\begin{table*}
\centering
\caption{Three convergence possibilities of the TM.}
\label{table:absorbingstates}
\begin{tabular}[]{|l|l|l|l|}
\hline
$(\mathrm{TA}_1,\mathrm{TA}_2)$& $s$ in general case & $s$ when $c=0.5$ & $a,b$\\
\hline
\hline

$(\mathrm{TA}_1,\mathrm{TA}_2)\rightarrow (E(x), E(\neg x))$ 
& $1<s<\min(s_1,s_4)$ 
& $1<s<\min(\frac{a+b}{a+1-b},\frac{a+b}{b+1-a})$
& $a>0.5, b>0.5$
\\
\hline

$(\mathrm{TA}_1,\mathrm{TA}_2)\rightarrow (I(x), E(\neg x))$ 
& $s_2<s<s_6$ 
& $\frac{a+b}{a}<s<\frac{a+b}{1-a}$
& $a>0.5$
\\
\hline

$(\mathrm{TA}_1,\mathrm{TA}_2)\rightarrow (E(x), I(\neg x))$ 
& $s_5<s<s_3$ 
& $\frac{a+b}{b}<s<\frac{a+b}{1-b}$
& $b>0.5$
\\
\hline

\end{tabular}
\end{table*}

%
%
%
%
\begin{figure}[htbp]
\begin{center}
\begin{minipage}{0.5\textwidth}
1. When $a>b>0.5$, the positions can be:\\

\begin{tikzpicture}[node distance = .35cm, font=\Huge]
    \tikzstyle{every node}=[scale=0.35]
        \node[label=below:$s_1$] at (1,4.3){};
    \node[label=below:$s_4$] at (2,4.3){};
    \node[label=below:$s_2$] at (4,4.3){};
    \node[label=below:$s_5$] at (5,4.3){};
    \node[label=below:$s_3$] at (6,4.3){};
    \node[label=below:$s_6$] at (7,4.3){};
    \node[label=below:$1$] at (0,4.3){};
    \node[label=below:$\infty$] at (8,4.3){};
    
    \draw [-{stealth[length=4mm]}] (0,4.38) -- (8,4.38);
    
    \draw[densely dashed] (1,4.4) -- (1,4.5);
    \draw[densely dashed] (2,4.4) -- (2,4.5);
    \draw[densely dashed] (4,4.4) -- (4,4.5);
    \draw[densely dashed] (5,4.4) -- (5,4.5);
    \draw[densely dashed] (6,4.4) -- (6,4.5);
    \draw[densely dashed] (7,4.4) -- (7,4.5);
\end{tikzpicture}

or
\vspace{0.2cm}

\begin{tikzpicture}[node distance = .35cm, font=\Huge]
    \tikzstyle{every node}=[scale=0.35]
        \node[label=below:$s_1$] at (1,4.3){};
    \node[label=below:$s_2$] at (4,4.3){};
    \node[label=below:$s_4$] at (4.5,4.3){};
    \node[label=below:$s_5$] at (5,4.3){};
    \node[label=below:$s_3$] at (6,4.3){};
    \node[label=below:$s_6$] at (7,4.3){};
    \node[label=below:$1$] at (0,4.3){};
    \node[label=below:$\infty$] at (8,4.3){};
    
    \draw [-{stealth[length=4mm]}] (0,4.38) -- (8,4.38);
    
    \draw[densely dashed] (1,4.4) -- (1,4.5);
    \draw[densely dashed] (4.5,4.4) -- (4.5,4.5);
    \draw[densely dashed] (4,4.4) -- (4,4.5);
    \draw[densely dashed] (5,4.4) -- (5,4.5);
    \draw[densely dashed] (6,4.4) -- (6,4.5);
    \draw[densely dashed] (7,4.4) -- (7,4.5);
\end{tikzpicture}
\end{minipage}

\vspace{0.5cm}
\begin{minipage}{0.5\textwidth}
2. When $b>a>0.5$, the positions can be:\\

\begin{tikzpicture}[node distance = .35cm, font=\Huge]
    \tikzstyle{every node}=[scale=0.35]
        \node[label=below:$s_4$] at (1,4.3){};
    \node[label=below:$s_1$] at (2,4.3){};
    \node[label=below:$s_5$] at (4,4.3){};
    \node[label=below:$s_2$] at (5,4.3){};
    \node[label=below:$s_6$] at (6,4.3){};
    \node[label=below:$s_3$] at (7,4.3){};
    \node[label=below:$1$] at (0,4.3){};
    \node[label=below:$\infty$] at (8,4.3){};
    
    \draw [-{stealth[length=4mm]}] (0,4.38) -- (8,4.38);
    
    \draw[densely dashed] (1,4.4) -- (1,4.5);
    \draw[densely dashed] (2,4.4) -- (2,4.5);
    \draw[densely dashed] (4,4.4) -- (4,4.5);
    \draw[densely dashed] (5,4.4) -- (5,4.5);
    \draw[densely dashed] (6,4.4) -- (6,4.5);
    \draw[densely dashed] (7,4.4) -- (7,4.5);
\end{tikzpicture}

or
\vspace{0.2cm}

\begin{tikzpicture}[node distance = .35cm, font=\Huge]
    \tikzstyle{every node}=[scale=0.35]
        \node[label=below:$s_4$] at (1,4.3){};
    \node[label=below:$s_5$] at (4,4.3){};
    \node[label=below:$s_1$] at (4.5,4.3){};
    \node[label=below:$s_2$] at (5,4.3){};
    \node[label=below:$s_6$] at (6,4.3){};
    \node[label=below:$s_3$] at (7,4.3){};
    \node[label=below:$1$] at (0,4.3){};
    \node[label=below:$\infty$] at (8,4.3){};
    
    \draw [-{stealth[length=4mm]}] (0,4.38) -- (8,4.38);
    
    \draw[densely dashed] (1,4.4) -- (1,4.5);
    \draw[densely dashed] (4.5,4.4) -- (4.5,4.5);
    \draw[densely dashed] (4,4.4) -- (4,4.5);
    \draw[densely dashed] (5,4.4) -- (5,4.5);
    \draw[densely dashed] (6,4.4) -- (6,4.5);
    \draw[densely dashed] (7,4.4) -- (7,4.5);
\end{tikzpicture}
\end{minipage}
\end{center}
\caption{\label{fig:positions} Positions of $s_1, ~s_2, ~s_3, ~s_4, ~s_5, ~s_6$ when $a>b>0.5$ or $b>a>0.5$, with $c \in (0, 1)$.}
\end{figure}
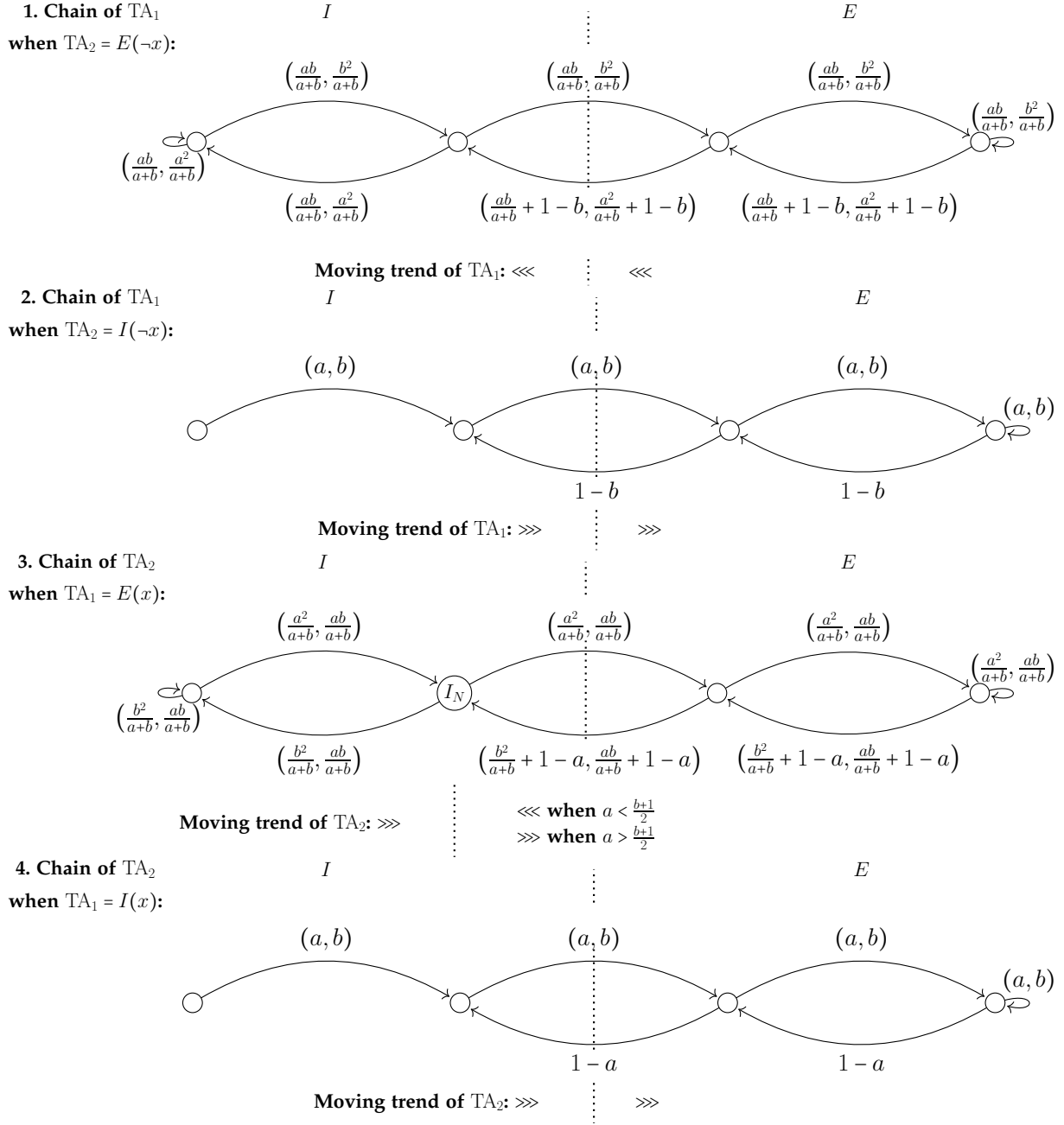
\begin{figure*}[htbp]
\begin{center}
\begin{minipage}{1\textwidth}
\resizebox{1\textwidth}{!}{
\begin{tikzpicture}[node distance = .35cm, font=\Huge]
    \tikzstyle{every node}=[scale=0.35]
    
    \node[state] (A) at (1,1) {};
    \node[state] (B) at (5,1) {};
    \node[state] (C) at (9,1) {};
    \node[state] (D) at (13,1) {};
    
    \node[thick] at (3,3) {$I$};
    \node[thick] at (11,3) {$E$};
    \node[thick] at (-0.6,3) {\bf 1. Chain of $\mathrm{TA}_1$};
    \node[thick] at (-0.6,2.5) {\bf when $\mathrm{TA}_2=E(\neg x)$:};
    \node[thick] at (4.5,-1) {\bf Moving trend of $\mathrm{TA}_1$: $\lll$};
    \node[thick] at (7.8,-1) {\bf $\lll$};

    \draw[dotted, thick] (7,0.3) -- (7,1.8);
    \draw[dotted, thick] (7,2.5) -- (7,3);
    \draw[dotted, thick] (7,-0.8) -- (7,-1.2);
    
    \draw[every loop]
    (A) edge[loop left] node [scale=1.2, below=0.1 of H] {$\left(\frac{ab}{a+b}, \frac{a^2}{a+b}\right)$} (A)
    (B) edge[bend left] node [scale=1.2, below=0.1 of H] {$\left(\frac{ab}{a+b}, \frac{a^2}{a+b}\right)$} (A) 
    (C) edge[bend left] node [scale=1.2, below=0.1 of H] {$\left(\frac{ab}{a+b}+1-b, \frac{a^2}{a+b}+1-b\right)$} (B) 
    (D) edge[bend left] node [scale=1.2, below=0.1 of C] {$\left(\frac{ab}{a+b}+1-b, \frac{a^2}{a+b}+1-b\right)$} (C);
    
     \draw[every loop]
    (A) edge[bend left] node [scale=1.2, above=0.1 of H] {$\left(\frac{ab}{a+b}, \frac{b^2}{a+b}\right)$} (B)
    (B) edge[bend left] node [scale=1.2, above=0.1 of H] {$\left(\frac{ab}{a+b}, \frac{b^2}{a+b}\right)$} (C) 
    (C) edge[bend left] node [scale=1.2, above=0.1 of H] {$\left(\frac{ab}{a+b}, \frac{b^2}{a+b}\right)$} (D) 
    (D) edge[loop right] node [scale=1.2, above=0.1 of C] {$\left(\frac{ab}{a+b}, \frac{b^2}{a+b}\right)$} (D);
\end{tikzpicture}
}

\resizebox{1\textwidth}{!}{
\begin{tikzpicture}[node distance = .35cm, font=\Huge]
    \tikzstyle{every node}=[scale=0.35]
    
    \node[state] (A) at (1,1) {};
    \node[state] (B) at (5,1) {};
    \node[state] (C) at (9,1) {};
    \node[state] (D) at (13,1) {};
    
    \node[thick] at (3,3) {$I$};
    \node[thick] at (11,3) {$E$};
    \node[thick] at (-0.6,3) {\bf 2. Chain of $\mathrm{TA}_1$};
    \node[thick] at (-0.6,2.5) {\bf when $\mathrm{TA}_2=I(\neg x)$:};
    \node[thick] at (4.5,-0.5) {\bf Moving trend of $\mathrm{TA}_1$: $\ggg$};
    \node[thick] at (7.8,-0.5) {\bf $\ggg$};

    \draw[dotted, thick] (7,0.3) -- (7,1.8);
    \draw[dotted, thick] (7,2.5) -- (7,3);
    \draw[dotted, thick] (7,-0.2) -- (7,-0.8);
    
    \draw[every loop]
    (C) edge[bend left] node [scale=1.2, below=0.1 of H] {$1-b$} (B) 
    (D) edge[bend left] node [scale=1.2, below=0.1 of C] {$1-b$} (C);
    
     \draw[every loop]
    (A) edge[bend left] node [scale=1.2, above=0.1 of H] {$(a,b)$} (B)
    (B) edge[bend left] node [scale=1.2, above=0.1 of H] {$(a,b)$} (C) 
    (C) edge[bend left] node [scale=1.2, above=0.1 of H] {$(a,b)$} (D) 
    (D) edge[loop right] node [scale=1.2, above=0.1 of C] {$(a,b)$} (D);

\end{tikzpicture}
}

\resizebox{1\textwidth}{!}{
\begin{tikzpicture}[node distance = .35cm, font=\Huge]
    \tikzstyle{every node}=[scale=0.35]
    
    \node[state] (A) at (1,1) {};
    \node[state] (B) at (5,1) {$I_N$};
    \node[state] (C) at (9,1) {};
    \node[state] (D) at (13,1) {};
    
    \node[thick] at (3,3) {$I$};
    \node[thick] at (11,3) {$E$};
    \node[thick] at (-0.6,3) {\bf 3. Chain of $\mathrm{TA}_2$};
    \node[thick] at (-0.6,2.5) {\bf when $\mathrm{TA}_1=E(x)$:};
    \node[thick] at (2.5,-1) {\bf Moving trend of $\mathrm{TA}_2$: $\ggg$};
    \node[thick] at (7,-0.8) {\bf $\lll$ when $a<\frac{b+1}{2}$};
    \node[thick] at (7,-1.2) {\bf $\ggg$ when $a>\frac{b+1}{2}$};

    \draw[dotted, thick] (7,0.3) -- (7,1.8);
    \draw[dotted, thick] (7,2.5) -- (7,3);
    \draw[dotted, thick] (5,-0.4) -- (5,-1.5);
    
    \draw[every loop]
    (A) edge[loop left] node [scale=1.2, below=0.1 of H] {$\left(\frac{b^2}{a+b}, \frac{ab}{a+b}\right)$} (A)
    (B) edge[bend left] node [scale=1.2, below=0.1 of H] {$\left(\frac{b^2}{a+b}, \frac{ab}{a+b}\right)$} (A) 
    (C) edge[bend left] node [scale=1.2, below=0.1 of H] {$\left(\frac{b^2}{a+b}+1-a, \frac{ab}{a+b}+1-a\right)$} (B) 
    (D) edge[bend left] node [scale=1.2, below=0.1 of C] {$\left(\frac{b^2}{a+b}+1-a, \frac{ab}{a+b}+1-a\right)$} (C);
    
     \draw[every loop]
    (A) edge[bend left] node [scale=1.2, above=0.1 of H] {$\left(\frac{a^2}{a+b}, \frac{ab}{a+b}\right)$} (B)
    (B) edge[bend left] node [scale=1.2, above=0.1 of H] {$\left(\frac{a^2}{a+b}, \frac{ab}{a+b}\right)$} (C) 
    (C) edge[bend left] node [scale=1.2, above=0.1 of H] {$\left(\frac{a^2}{a+b}, \frac{ab}{a+b}\right)$} (D) 
    (D) edge[loop right] node [scale=1.2, above=0.1 of C] {$\left(\frac{a^2}{a+b}, \frac{ab}{a+b}\right)$} (D);
\end{tikzpicture}
}

\resizebox{1\textwidth}{!}{
\begin{tikzpicture}[node distance = .35cm, font=\Huge]
    \tikzstyle{every node}=[scale=0.35]
    
    \node[state] (A) at (1,1) {};
    \node[state] (B) at (5,1) {};
    \node[state] (C) at (9,1) {};
    \node[state] (D) at (13,1) {};
    
    \node[thick] at (3,3) {$I$};
    \node[thick] at (11,3) {$E$};
    \node[thick] at (-0.6,3) {\bf 4. Chain of $\mathrm{TA}_2$};
    \node[thick] at (-0.6,2.5) {\bf when $\mathrm{TA}_1=I(x)$:};
    \node[thick] at (4.5,-0.5) {\bf Moving trend of $\mathrm{TA}_2$: $\ggg$};
    \node[thick] at (7.8,-0.5) {\bf $\ggg$};

    \draw[dotted, thick] (7,0.3) -- (7,1.8);
    \draw[dotted, thick] (7,2.5) -- (7,3);
    \draw[dotted, thick] (7,-0.2) -- (7,-0.8);
    
    \draw[every loop]
    (C) edge[bend left] node [scale=1.2, below=0.1 of H] {$1-a$} (B) 
    (D) edge[bend left] node [scale=1.2, below=0.1 of C] {$1-a$} (C);
    
     \draw[every loop]
    (A) edge[bend left] node [scale=1.2, above=0.1 of H] {$(a,b)$} (B)
    (B) edge[bend left] node [scale=1.2, above=0.1 of H] {$(a,b)$} (C) 
    (C) edge[bend left] node [scale=1.2, above=0.1 of H] {$(a,b)$} (D) 
    (D) edge[loop right] node [scale=1.2, above=0.1 of C] {$(a,b)$} (D);

\end{tikzpicture}
}
\end{minipage}
\end{center}
\caption{\label{fig:trend} Moving trend when $s\in (s_2, s_5)$. In the noisy case, where $ P(y=1 | X=1) = a$, and $P(y=1 | X=0) = b$; $P(X=1)=c$; $0.5<b<a<1, ~c \in (0,1)$.}
\end{figure*}

Three convergence possibilities of the TM are summarized in Table \ref{table:absorbingstates}. This table provides important insights, based on which we analyze how $a$, $b$, $c$, and $s$ affect the convergence of the TM. We organize our analysis according to the following two cases, unbiased and biased training data.

{\bf Unbiased data}.
We first simplify the analysis by rendering $c=0.5$. This means training samples are unbiased. The conditions for $s$ are listed in the third column in Table \ref{table:absorbingstates}.

\begin{enumerate}
\item When $a<0.5, b>0.5$, the NOT relation is dominant in the training data.\\
In this case, $s_4=\min(s_1,s_4)<1$, $s_6<s_2$, which means the  intervals $(1, \min(s_1,s_4))$ and $(s_2, s_6)$ do not exist. Therefore, only when $s\in (s_5,s_3)$, the TM can converge to $(E(x),I(\neg x))$, reflecting the NOT relation.

\item When $a>0.5, b<0.5$, IDENTITY is the dominant relation in the training data.\\
In this case, $s_1=\min(s_1,s_4)<1$, $s_3<s_5$, the TM converges only when $s\in (s_2,s_6)$, to $(I(x),E(\neg x))$, i.e., the IDENTITY relation.

\item When $a>0.5, b>0.5$, both IDENTITY and NOT are dominating in the training data. Figure \ref{fig:positions} shows the positions of $s_1, ..., s_6$ when $a>b>0.5$ and $b>a>0.5$ respectively.
\begin{enumerate}
    
    \item {\color{black}When $a>b>0.5$, we have $s_4<s_5<s_3$ and $s_4<s_6$, implying two possible positions of $s_4$ that are visualized by item 1 in Fig.~\ref{fig:positions}. Similarly, when $b>a>0.5$, we have $s_1<s_2<s_6$ and $s_1<s_3$. Item 2 in Fig.~\ref{fig:positions} depicts the possible positions of $s_1$. Indeed, Fig.~\ref{fig:positions} shows that $s_1$ and $s_4$ cannot be greater than $\min(s_2,s_3,s_5,s_6)$ at the same time, i.e., $\min(s_1,s_4)<\min(s_2,s_3,s_5,s_6)$.} Therefore, when $s \in (1,\min(s_1,s_4))$, the TM converges only to $(E(x), E(\neg x))$. In other words, the $s$ satisfying the condition for converging to $(E(x), E(\neg x))$ will not result in the convergence to $(I(x), E(\neg x))$ or $(E(x), I(\neg x))$.
    
    In the classifying process, $C(x) = E(x) \wedge E(\neg x)$ means all inputs, whether $x=0$ or $x=1$, can be classified into pattern $y=1$, reflecting the fact that both IDENTITY and NOT are the sub-patterns associated with pattern $y=1$ in this case.
    
    \item When $a>b>0.5$, Figure \ref{fig:positions} shows that the interval of $(s_2,s_6)$ contains the interval of $(s_5,s_3)$. This means:
    \begin{itemize}
        \item When $s\in (s_2, s_5) \bigcup (s_3,s_6)$, the TM will converge to $(I(x),E(\neg x))$.
        \item When $s\in (s_5, s_3)$, the TM can converge to either $(I(x),E(\neg x))$ or $(E(x),I(\neg x))$. 
    \end{itemize}
    We take the interval $s\in (s_2, s_5)$ as an example to show how the TM converges to $(I(x),E(\neg x))$.
    
    When $s$ is bounded by $(s_2, s_5)$, the transition probabilities in the Markov chains are bounded accordingly, as shown in Figure \ref{fig:trend}. The ranges of the transition probabilities imply the trend of movement of the TAs. In Chain $1$, the trend is $\mathrm{TA}_1$ moving towards $I(x)$. In Chain $2$ and Chain $4$, the trend is that $\mathrm{TA}_1$ and $\mathrm{TA}_2$ both move towards excluding literal. The situation in Chain $3$ is different. When $\mathrm{TA}_2$ is on the left-hand side of the chain, the trend is that it moves towards $E(\neg x)$. However, when $\mathrm{TA}_2$ is on the right-hand side of the chain, the trend depends on the values of $a$ and $b$. In brief, if $a>\frac{b+1}{2}$, $\mathrm{TA}_2$ tends to move towards $E(\neg x)$, whereas if $a<\frac{b+1}{2}$, $\mathrm{TA}_2$ tends to move towards $I(\neg x)$.
    
    Suppose both the TAs initially are in one of the include literal-states, i.e., $(\mathrm{TA}_1, \mathrm{TA}_2) = (I(x), I(\neg x))$. Chain $2$ and Chain $4$ tell that both TAs tend to move towards $E(\cdot)$. 
    \begin{itemize}
        \item If $\mathrm{TA}_2$ reaches $E(\neg x)$ first, then Chain $1$ shows that $\mathrm{TA}_1$ tends to move towards $I(x)$. Once $\mathrm{TA}_1$ reaches $I(x)$, according to Chain~$4$, $\mathrm{TA}_2$ tends to move towards $E(\neg x)$. The TM thus converges to $(I(x), E(\neg x))$.
        
        \item If $\mathrm{TA}_1$ reaches $E(x)$ first, then Chain $3$ shows that $\mathrm{TA}_2$ tends to first move towards $E(\neg x)$. When $\mathrm{TA}_2$ reaches the node $I_N$, that is in the middle left of the chain, $\mathrm{TA}_2$ tends to move towards either $I(\neg x)$ or $E(\neg x)$: 
        \begin{itemize}
            \item when $a>\frac{b+1}{2}$, $\mathrm{TA}_2$ continues the trend of moving towards $E(\neg x)$. Once $\mathrm{TA}_2$ reaches the state of $E(\neg x)$, $\mathrm{TA}_1$ tends to move towards $I(x)$ according to Chain $1$. This means the TM is moving towards $(I(x), E(\neg x))$, and once $\mathrm{TA}_1=I(x)$ and $\mathrm{TA}_2=E(\neg x)$, the TAs tend to go deeper and deeper towards the end states and the TM converges to $(I(x), E(\neg x))$.
            \item when $a<\frac{b+1}{2}$, the probability that $\mathrm{TA}_2$ moves towards $E(\neg x)$ is less than the probability it moves towards $I(\neg x)$. Therefore, $\mathrm{TA}_2$ tends to linger around at the node $I_N$. When it happens that $\mathrm{TA}_2$ moves towards $E(\neg x)$ and reaches the node that is on the right-hand side of the chain, then according to Chain $1$, $\mathrm{TA}_1$ tends to move towards $I(x)$. This means that it is still possible for the TM to move towards $(I(x), E(\neg x))$, but the process that the TM reaches $(I(x), E(\neg x))$ involves more ``back and forth'' movements, and thus it takes longer time for the TM to converge. This makes sense, as when $a<\frac{b+1}{2}$, the difference between $a$ and $b$ is smaller than when $a>\frac{b+1}{2}$, it is reasonable that the TM will need more time to converge to the more advantageous relation.
        \end{itemize}
    \end{itemize}
    Indeed, depending on the range of $s$, the TM converges through different paths. One can examine how different choices of $s$ influence the converging process by following the same Markov analysis on the above example.
    
    
    \item When $b>a>0.5$, the situation is completely opposite to the case when $a>b>0.5$. Figure \ref{fig:positions} shows the interval of $(s_2,s_6)$ is contained in $(s_5,s_3)$.
    \begin{itemize}
        \item When $s \in (s_5,s_2)\bigcup (s_6,s_3)$, the TM converges only to $(E(x),I(\neg x))$, i.e., the NOT operator.
        \item When $s \in (s_2,s_6)$, the TM converges to either $(I(x),E(\neg x))$ or $(E(x),I(\neg x))$. 
    \end{itemize}
\end{enumerate}

When $a>0.5, b>0.5$, there are basically two sub-patterns (IDENTITY and NOT) in the pattern $y=1$. By properly configuring $s$, a TM with only one clause is able to converge to $(E(x), E(\neg x))$ to classify both sub-patterns into the correct pattern $y=1$. However, the one-clause TM does not distinguish between sub-patterns. To distinguish between sub-patterns, we need to configure the TM with at least two clauses, each of which being used to represent one sub-pattern. These clauses can be organized in a voting architecture, where the threshold $T=1$ can be applied to allocate clause resource.

\item When $a=b=0.5$, we have $s_1=s_4=1$, and $s_2=s_6=s_5=s_3=2$. No $s$-value will cause the TM to converge. This makes sense as in this case, the training samples have the greatest uncertainty, there are basically no pattern present for  the TM to learn.

\item When $a<0.5, ~b<0.5$, we have $s_1<1$, $s_4<1$, $s_6<s_2$, and $s_3<s_5$. No $s$-value meets the convergence conditions listed in Table \ref{table:absorbingstates}. The TM accordingly does not learn, because it is designed to learn the pattern with label $y=1$. For this scenario, $y=1$ is not the dominant pattern in the training set because both $a$ and $b$ are less than $0.5$. 
Indeed, the dominant pattern becomes $y=0$. So if we swap Type I and Type II Feedback for the TM, it is able to learn the pattern $y=0$, with the analysis being exactly the same as that on the case when $a>0.5, b>0.5$ for pattern $y=1$.
\end{enumerate}

\begin{figure*}[htbp]
\begin{center}
\begin{minipage}{0.6\textwidth}
{\bf The intervals of $c$:}\\

\begin{tikzpicture}[node distance = .35cm, font=\Huge]
    \tikzstyle{every node}=[scale=0.35]
        \node[label=below:$c_1$] at (2,4.3){};
    \node[label=below:$c_2$] at (4,4.3){};
    \node[label=below:$c_3$] at (6,4.3){};
    \node[label=below:$c_4$] at (8,4.3){};
    \node[label=below:$0$] at (0,4.3){};
    \node[label=below:$1$] at (10,4.3){};
    
    \draw  (0,4.38) -- (10,4.38);
    \draw[densely dashed] (0,4.4) -- (0,4.5);
    \draw[densely dashed] (2,4.4) -- (2,4.5);
    \draw[densely dashed] (4,4.4) -- (4,4.5);
    \draw[densely dashed] (6,4.4) -- (6,4.5);
    \draw[densely dashed] (8,4.4) -- (8,4.5);
    \draw[densely dashed] (10,4.4) -- (10,4.5);
\end{tikzpicture}
\end{minipage}
\begin{minipage}{0.6\textwidth}
{\bf The positions of $(s_2, s_6)$ and $(s_5, s_3)$:}\\
\end{minipage}
\begin{minipage}{0.6\textwidth}
1. When $c<c_1$:\\
\begin{tikzpicture}[font=\tiny\sffamily]
    \node[label=below:$s_5$] at (1,4.5){};
    \node[label=below:$s_3$] at (2,4.5){};
    \node[label=below:$s_2$] at (5,4.5){};
    \node[label=below:$s_6$] at (9,4.5){};
    \node[label=below:$1$] at (0,4.5){};
    \node[label=below:$\infty$] at (10,4.5){};
    
    \draw (0,4.38) -- (3,4.38);
    \draw[densely dashed] (3,4.38) -- (4,4.38);
    \draw [-{stealth[length=4mm]}] (4,4.38) -- (10,4.38);
    
    \draw (1, 4.38) -- (1, 4.88) -- (2, 4.88) -- (2, 4.38);
    \draw (5, 4.38) -- (5, 4.98) -- (9, 4.98) -- (9, 4.38);  
\end{tikzpicture}
\end{minipage}
\begin{minipage}{0.6\textwidth}
2. When $c=c_1$:\\
\begin{tikzpicture}[font=\tiny\sffamily]
    \node[label=below:$s_5$] at (3,4.5){};
    \node[label=below:$s_3$] at (4.3,4.5){};
    \node[label=below:$s_2$] at (4.7,4.5){};
    \node[label=below:$s_6$] at (8.75,4.5){};
    \node[label=below:$1$] at (0,4.5){};
    \node[label=below:$\infty$] at (10,4.5){};
    
    \draw [-{stealth[length=4mm]}] (0,4.38) -- (10,4.38);
    
    \draw (3, 4.38) -- (3, 4.88) -- (4.5, 4.88) -- (4.5, 4.38);
    \draw (4.5, 4.38) -- (4.5, 4.98) -- (8.75, 4.98) -- (8.75, 4.38);
\end{tikzpicture}
\end{minipage}
\begin{minipage}{0.6\textwidth}
3. When $c_1<c<c_2$:\\
\begin{tikzpicture}[font=\tiny\sffamily]
    \node[label=below:$s_5$] at (3.25,4.5){};
    \node[label=below:$s_3$] at (5,4.5){};
    \node[label=below:$s_2$] at (4.25,4.5){};
    \node[label=below:$s_6$] at (8.25,4.5){};
    \node[label=below:$1$] at (0,4.5){};
    \node[label=below:$\infty$] at (10,4.5){};
    
    \draw [-{stealth[length=4mm]}] (0,4.38) -- (10,4.38);
    
    \draw (3.25, 4.38) -- (3.25, 4.88) -- (5, 4.88) -- (5, 4.38);
    \draw (4.25, 4.38) -- (4.25, 4.98) -- (8.25, 4.98) -- (8.25, 4.38);
\end{tikzpicture}
\end{minipage}
\begin{minipage}{0.6\textwidth}
4. When $c=c_2$:\\
\begin{tikzpicture}[font=\tiny\sffamily]
    \node[label=below:$s_5$] at (3.8,4.5){};
    \node[label=below:$s_3$] at (6,4.5){};
    \node[label=below:$s_2$] at (4.2,4.5){};
    \node[label=below:$s_6$] at (7.75,4.5){};
    \node[label=below:$1$] at (0,4.5){};
    \node[label=below:$\infty$] at (10,4.5){};
    
    \draw [-{stealth[length=4mm]}] (0,4.38) -- (10,4.38);
    
    \draw (4, 4.38) -- (4, 4.88) -- (6, 4.88) -- (6, 4.38);
    \draw (4, 4.38) -- (4, 4.98) -- (7.75, 4.98) -- (7.75, 4.38);
\end{tikzpicture}
\end{minipage}
\begin{minipage}{0.6\textwidth}
5. When $c_2<c<c_3$:\\
\begin{tikzpicture}[font=\tiny\sffamily]
    \node[label=below:$s_5$] at (4.25,4.5){};
    \node[label=below:$s_3$] at (6.5,4.5){};
    \node[label=below:$s_2$] at (3.5,4.5){};
    \node[label=below:$s_6$] at (7.25,4.5){};
    \node[label=below:$1$] at (0,4.5){};
    \node[label=below:$\infty$] at (10,4.5){};
    
    \draw [-{stealth[length=4mm]}] (0,4.38) -- (10,4.38);
    
    \draw (4.25, 4.38) -- (4.25, 4.88) -- (6.5, 4.88) -- (6.5, 4.38);
    \draw (3.5, 4.38) -- (3.5, 4.98) -- (7.25, 4.98) -- (7.25, 4.38);
\end{tikzpicture}
\end{minipage}
\begin{minipage}{0.6\textwidth}
6. When $c=c_3$:\\
\begin{tikzpicture}[font=\tiny\sffamily]
    \node[label=below:$s_5$] at (4.5,4.5){};
    \node[label=below:$s_3$] at (6.8,4.5){};
    \node[label=below:$s_2$] at (3.375,4.5){};
    \node[label=below:$s_6$] at (7.2,4.5){};
    \node[label=below:$1$] at (0,4.5){};
    \node[label=below:$\infty$] at (10,4.5){};
    
    \draw [-{stealth[length=4mm]}] (0,4.38) -- (10,4.38);
    
    \draw (4.5, 4.38) -- (4.5, 4.88) -- (7, 4.88) -- (7, 4.38);
    \draw (3.375, 4.38) -- (3.375, 4.98) -- (7, 4.98) -- (7, 4.38);
\end{tikzpicture}
\end{minipage}
\begin{minipage}{0.6\textwidth}
7. When $c_3<c<c_4$:\\
\begin{tikzpicture}[font=\tiny\sffamily]
    \node[label=below:$s_5$] at (5.25,4.5){};
    \node[label=below:$s_3$] at (8,4.5){};
    \node[label=below:$s_2$] at (2.75,4.5){};
    \node[label=below:$s_6$] at (6,4.5){};
    \node[label=below:$1$] at (0,4.5){};
    \node[label=below:$\infty$] at (10,4.5){};
    
    \draw [-{stealth[length=4mm]}] (0,4.38) -- (10,4.38);
    
    \draw (5.25, 4.38) -- (5.25, 4.88) -- (8, 4.88) -- (8, 4.38);
    \draw (2.75, 4.38) -- (2.75, 4.98) -- (6, 4.98) -- (6, 4.38);
\end{tikzpicture}
\end{minipage}
\begin{minipage}{0.6\textwidth}
8. When $c=c_4$:\\
\begin{tikzpicture}[font=\tiny\sffamily]
    \node[label=below:$s_5$] at (5.7,4.5){};
    \node[label=below:$s_3$] at (8.75,4.5){};
    \node[label=below:$s_2$] at (2.5,4.5){};
    \node[label=below:$s_6$] at (5.3,4.5){};
    \node[label=below:$1$] at (0,4.5){};
    \node[label=below:$\infty$] at (10,4.5){};
    
    \draw [-{stealth[length=4mm]}] (0,4.38) -- (10,4.38);
    
    \draw (5.5, 4.38) -- (5.5, 4.88) -- (8.75, 4.88) -- (8.75, 4.38);
    \draw (2.5, 4.38) -- (2.5, 4.98) -- (5.5, 4.98) -- (5.5, 4.38);
\end{tikzpicture}
\end{minipage}
\begin{minipage}{0.6\textwidth}
9. When $c>c_4$:\\
\begin{tikzpicture}[font=\tiny\sffamily]
    \node[label=below:$s_2$] at (1,4.5){};
    \node[label=below:$s_6$] at (2.5,4.5){};
    \node[label=below:$s_5$] at (5.5,4.5){};
    \node[label=below:$s_3$] at (9,4.5){};
    \node[label=below:$1$] at (0,4.5){};
    \node[label=below:$\infty$] at (10,4.5){};
    
    \draw (0,4.38) -- (3,4.38);
    \draw[densely dashed] (3,4.38) -- (4,4.38);
    \draw [-{stealth[length=4mm]}] (4,4.38) -- (10,4.38);
    
    \draw (1, 4.38) -- (1, 4.98) -- (2.5, 4.98) -- (2.5, 4.38);
    \draw (5.5, 4.38) -- (5.5, 4.88) -- (9, 4.88) -- (9, 4.38);  
\end{tikzpicture}
\end{minipage}
\end{center}
\caption{\label{fig:positionswithc} How the positions of $s_2, ~s_3, ~s_5, ~s_6$ change with $c$, in the environment $a>b>0.5$.}
\end{figure*}
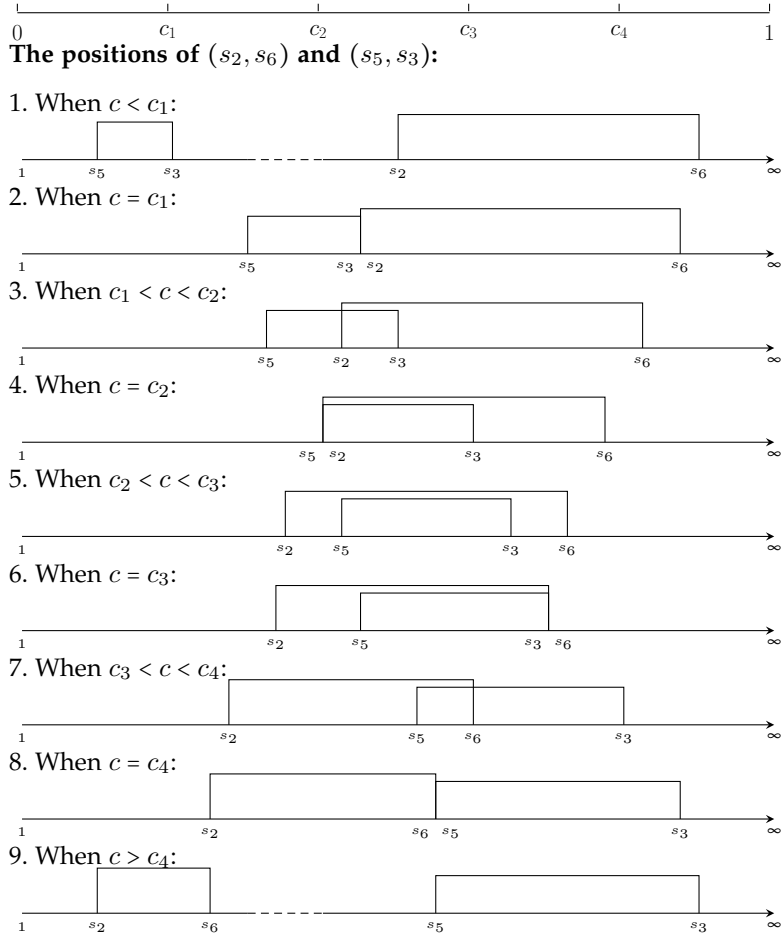

{\bf Biased data}. We now analyze the convergence of the TM in a more generic environment where training samples can be biased, i.e., when $0<c<1$ and $c\neq 0.5$. We state that the general case share the same conclusions with the unbiased case for the following environments: (1) $a<0.5, b>0.5$, (2) $a>0.5, b<0.5$, (4) $a=0.5, b=0.5$, and (5) $a<0.5, b<0.5$.
The difference lies in the environment where $a>0.5, b>0.5$. We omit the deductive process and summarize the analytical results for the environment where $a>0.5$ and $b>0.5$ below.

\begin{enumerate}
\item When $s < \min(s_1,s_4)$, the TM converges to $(E(x), E(\neg x))$ in both cases where $a>b>0.5$ and $b>a>0.5$. This part is the same as in the unbiased environment.

\item We then observe how the positions of $s_2$, $s_3$, $s_5$, and $s_6$ change as $c$ changes. We plot the changes in the case where $a>b>0.5$ in Figure~\ref{fig:positionswithc}. There we have defined 
{\footnotesize
\begin{align}
c_1 = \frac{1-b}{a+1-b},~ 
c_2 = \frac{b}{a+b}, ~
c_3 = \frac{1-b}{2-a-b}, ~
c_4 = \frac{b}{1-a+b}. \nonumber 
\end{align}
}

As can be seen from Figure \ref{fig:positionswithc}, the lengths and the positions of the intervals $(s_5, s_3)$ and $(s_2, s_6)$ depend on where $c$ is. Briefly speaking, the smaller the $c$, the shorter is the interval $(s_5, s_3)$, the longer is the interval $(s_2, s_6)$, and the farther away are these two intervals from each other. When $c \rightarrow 0$, we have $(s_2, s_6) \rightarrow (\infty, \infty)$, $(s_5, s_3) \rightarrow (1, \frac{b}{1-b})$, and the distance between these two intervals becomes the largest. When $c<c_1$, there is no overlap between intervals $(s_5, s_3)$ and $(s_2, s_6)$. 
When $c$ increases, the distance between these two intervals decreases, and when $c=c_1$, they become next to each other. If $c$ continues to increase and moves into $(c_1, c_2)$, the intervals $(s_5,s_3)$ and $(s_2,s_6)$ start overlapping each other. When $c_2 \leq c \leq c_3$, $(s_5,s_3)$ ``merges'' into $(s_2,s_6)$.

If $c$ continues on increasing, $(s_5,s_3)$ starts moving out of $(s_2,s_6)$. When $c>c_4$, $(s_5,s_3)$ and $(s_2,s_6)$ part from each other again and become farther away from each other. When $c \rightarrow 1$, we have $(s_5, s_3) \rightarrow (\infty, \infty)$, $(s_2, s_6) \rightarrow (1, \frac{a}{1-a})$, and the distance between $(s_5,s_3)$ and $(s_2,s_6)$ reaches the greatest again.

Wherever $c$ is, if $s\in (s_5, s_3)$, the TM converges to the NOT relation, i.e., $(\mathrm{TA}_1, \mathrm{TA}_2)=(E(x), I(\neg x))$; if $s\in (s_2, s_6)$, the TM converges to the IDENTITY relation, i.e., $(\mathrm{TA}_1, \mathrm{TA}_2)=(I(x), E(\neg x))$; and if $s$ is in the overlapping region, the TM converges to either $(I(x), E(\neg x))$ or $(E(x), I(\neg x))$.

Figure \ref{fig:positionswithc} indeed reveals a crucial property of TMs: \emph{by properly selecting a value for $s$, the TM is able to pick up the pattern represented by very rare samples}. For example, in the environment where $c$ is very small, most of the samples have input $x=0$, whose label is $y=1$ with probability $b>0.5$. Yet, there are very rare input $x=1$ samples, whose label is $y=1$ with probability $a>b>0.5$. In this environment, if we set $s\in(s_5, s_3)$, then the TM will converge to $(E(x), I(\neg x))$, capturing the NOT relation represented by the dominating samples $x=0$. However, if we configure $s$ to be sufficiently large, that is, within the interval $(s_2, s_6)$, then the TM will still converge to the IDENTITY relation $(I(x), E(\neg x))$, despite the low value of $c$.  This ability to directly control the frequency of the patterns captured is a unique property of TMs.
\end{enumerate}

\begin{remark}
\label{remark:onebitnoise}
The convergence of TM in the noisy case do have requirements on the depths of the TAs. The tolerance of noise in TM relies on the depth of states associated with each action. The  larger the depth, the more the robustness to noise.
\end{remark}

\section{Conclusions}\label{conclusion}
In this article, we studied the convergence of single-clause Tsetlin Machines (TMs) for learning two of the most fundamental logical operators, i.e., the IDENTITY and the NOT operators. The analytical results reveal that the TM, in its simplest form, can learn the above two propositional logic operators over an infinite training horizon. Besides, by appropriately configuring the granularity parameter $s$, the TM is able to capture the pattern represented by rare samples and select the most accurate one when two candidate patterns are incompatible. This ability to directly control the frequency and the accuracy of the patterns captured is a unique property of TMs.

The analytical approach proposed in this paper  provides a foundation for formal treatment of TM learning in more complex scenarios. To this end, in our further work, we aim to extend our analysis to also cover the binary logical operators: AND, OR, and XOR. Such an analysis will involve the voting mechanism of the TM in full extent.  

\section*{Acknowledgment}
This work is supported by the project Spacetime Vision: Towards Unsupervised Learning in the 4D World financed by the EEA and Norway Grants 2014-2021 under the grant number EEA-RO-NO-2018-04.

\bibliographystyle{IEEEtran}
\bibliography{references}

{\color{black}
\begin{IEEEbiography}[{\includegraphics[width=1.in,height=1.45in,clip,keepaspectratio]{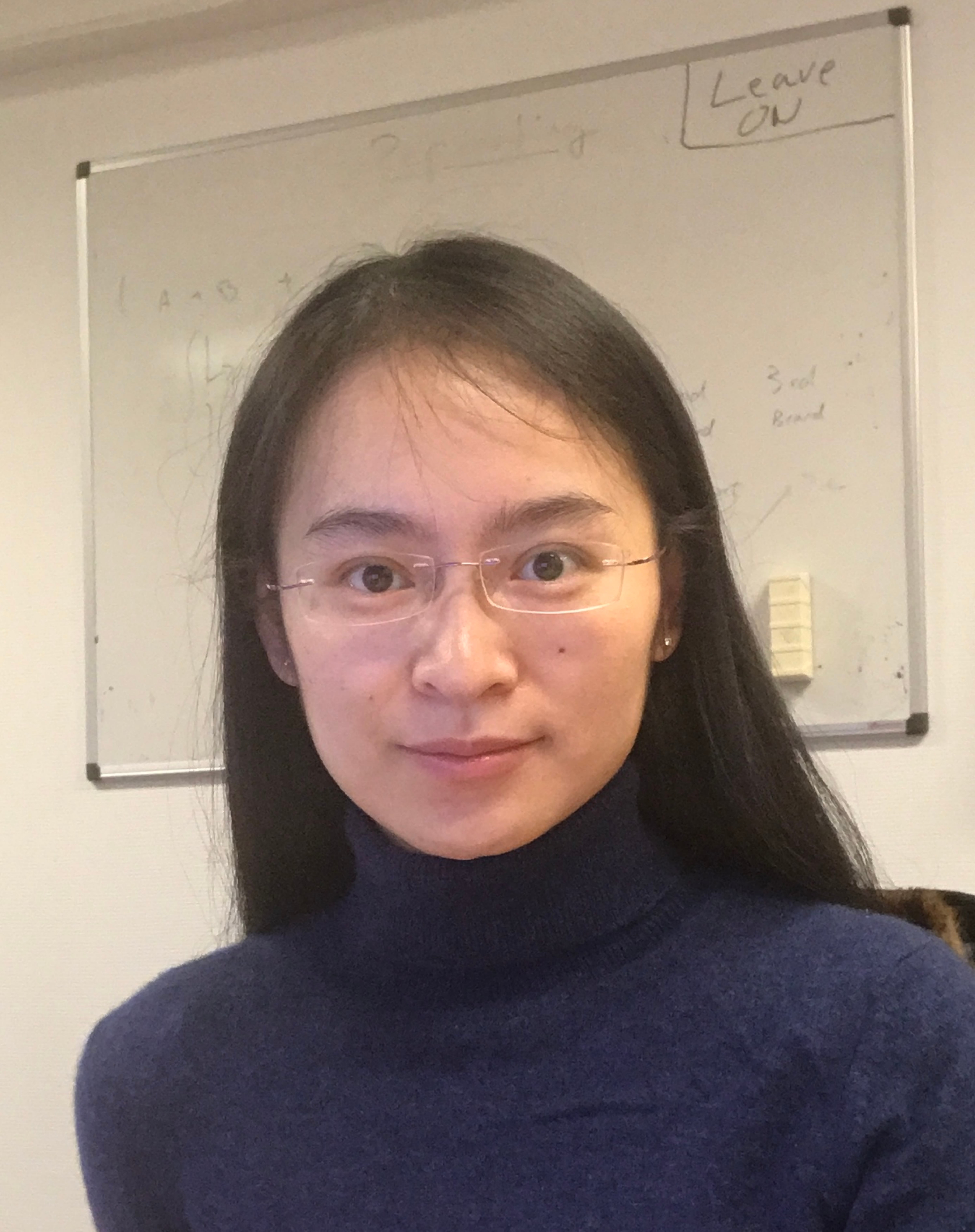}}]{Xuan Zhang}
received her PhD degree in Information and Communication Technology from
the University of Agder (UiA) in 2015. She has a Master's
degree in Signal and Information Processing and a Bachelor's degree in Electronics and Information Engineering. She is now a senior researcher in Norwegian Research Center (NORCE). At the same time, she is a scientific researcher
at Centre of Artificial Intelligence Research (CAIR) in UiA. She is currently serving as a Board Member of Norwegian Association for Image Processing and Machine Learning.  Her research interests include:
Learning Automata, Mathematical Analysis on Learning Algorithms, Stochastic Modeling and Optimization, Deep Learning, Natural Language Processing and Computer Vision.
\end{IEEEbiography}
}

\begin{IEEEbiography}[{\includegraphics[width=1.in,height=1.45in,clip,keepaspectratio]{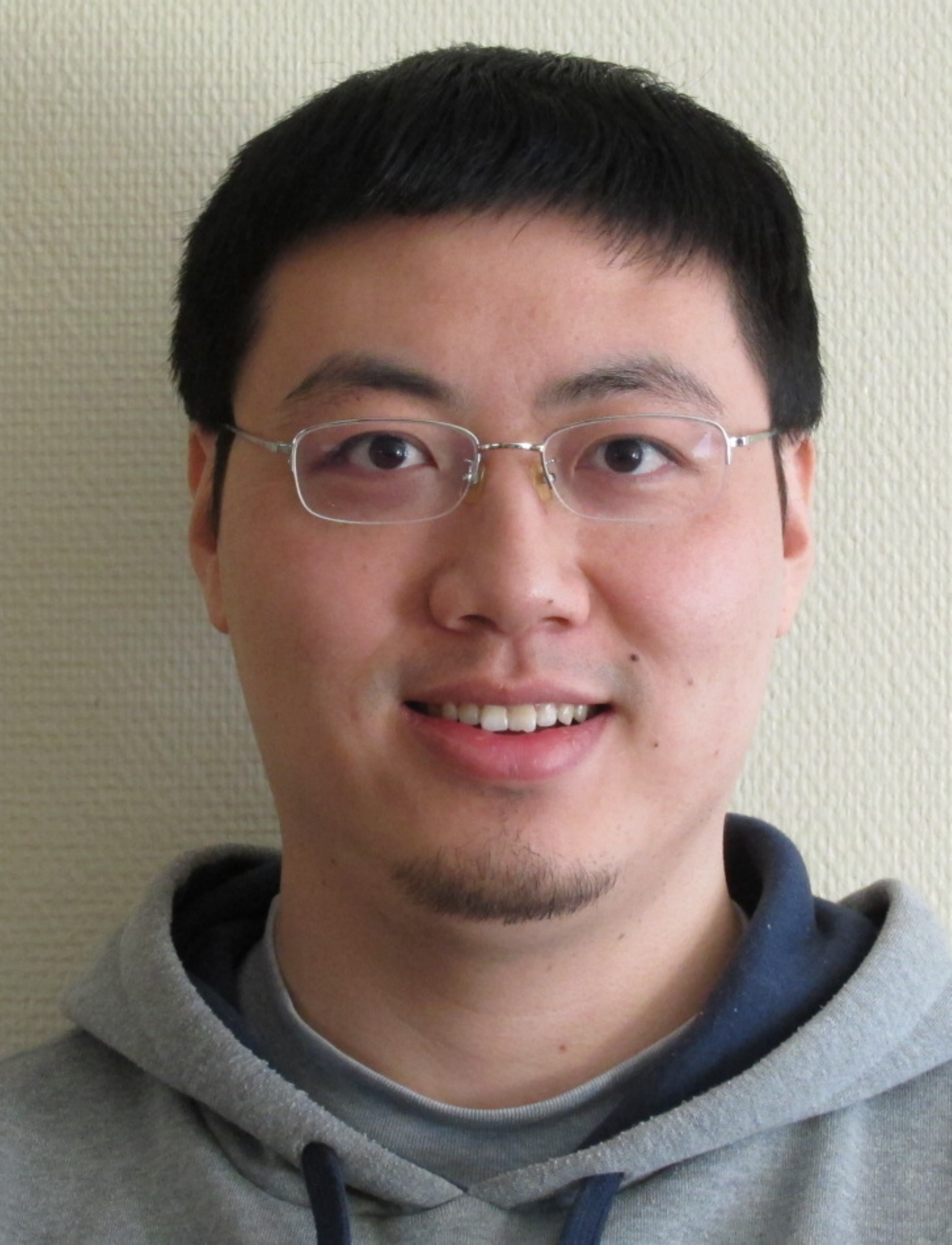}}]{Lei Jiao}
(M'12-SM'18) received his BE degree from Hunan University, China, in 2005. He received his ME degree from Shandong University, China, in 2008. He obtained his PhD degree in Information and  Communications Technology from University of Agder (UiA), Norway, in 2012. He is now an Associated Professor in the Department of Information and Communication Technology, UiA. His research interests include reinforcement learning, learning automata, natural language processing, resource allocation and performance evaluation for communication and energy systems.
\end{IEEEbiography}
\begin{IEEEbiography}[{\includegraphics[width=1.in,height=1.45in,clip,keepaspectratio]{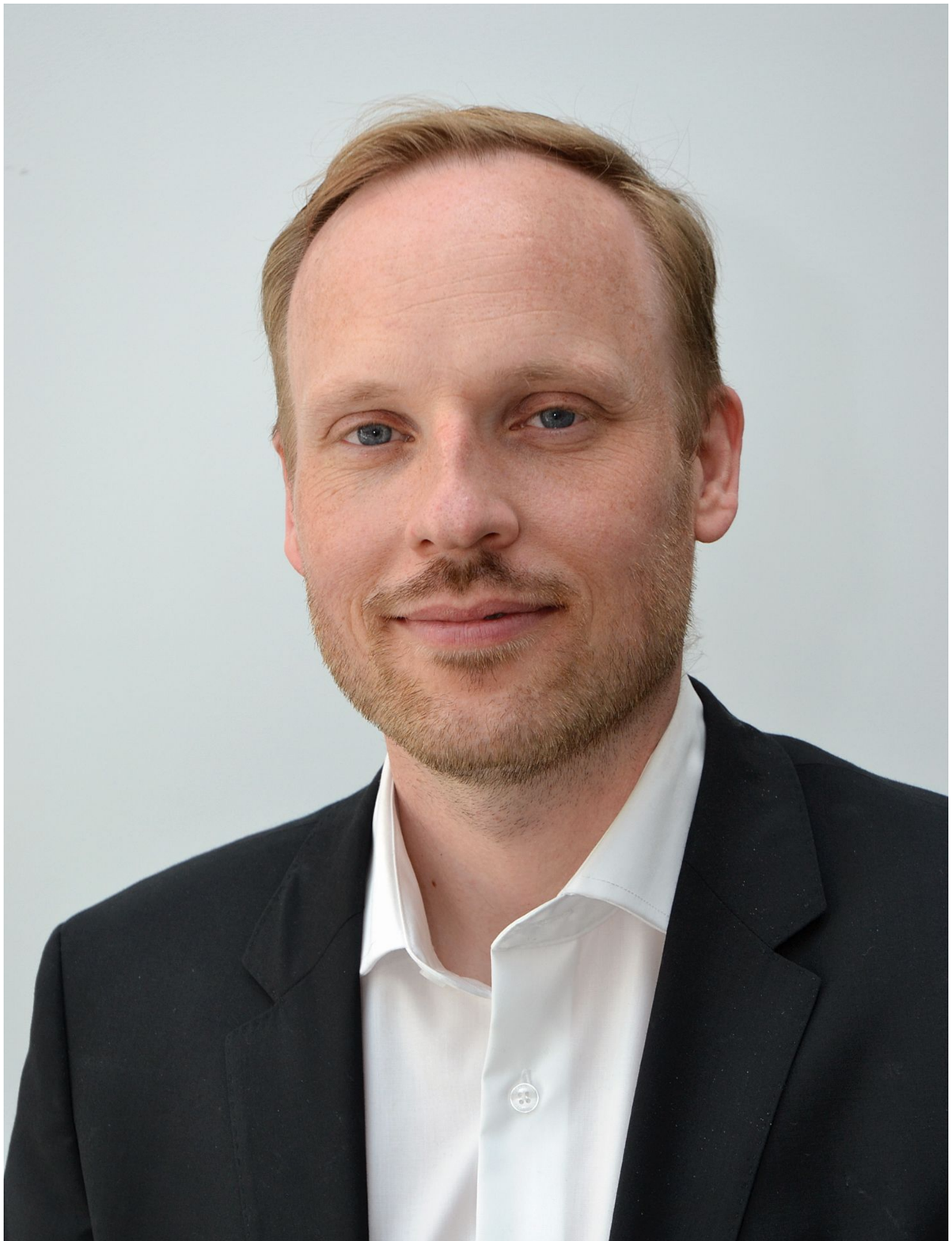}}]{Ole-Christoffer Granmo} is a Professor and Founding Director of Centre for Artificial Intelligence Research (CAIR), University of Agder, Norway. He obtained his master’s degree in 1999 and the PhD degree in 2004, both from the University of Oslo, Norway. Dr. Granmo has authored in excess of 140 refereed papers with numerous best paper awards, encompassing learning automata, bandit algorithms, Tsetlin machines, Bayesian reasoning, reinforcement learning, and computational linguistics. He has further coordinated 7+ Norwegian Research Council projects and graduated more than 60 master- and PhD students. Dr. Granmo is also a co-founder of the Norwegian Artificial Intelligence Consortium (NORA). Apart from his academic endeavours, he co-founded Anzyz Technologies AS.. 
\end{IEEEbiography}
\begin{IEEEbiography}[{\includegraphics[width=1.in,height=1.4in,clip,keepaspectratio]{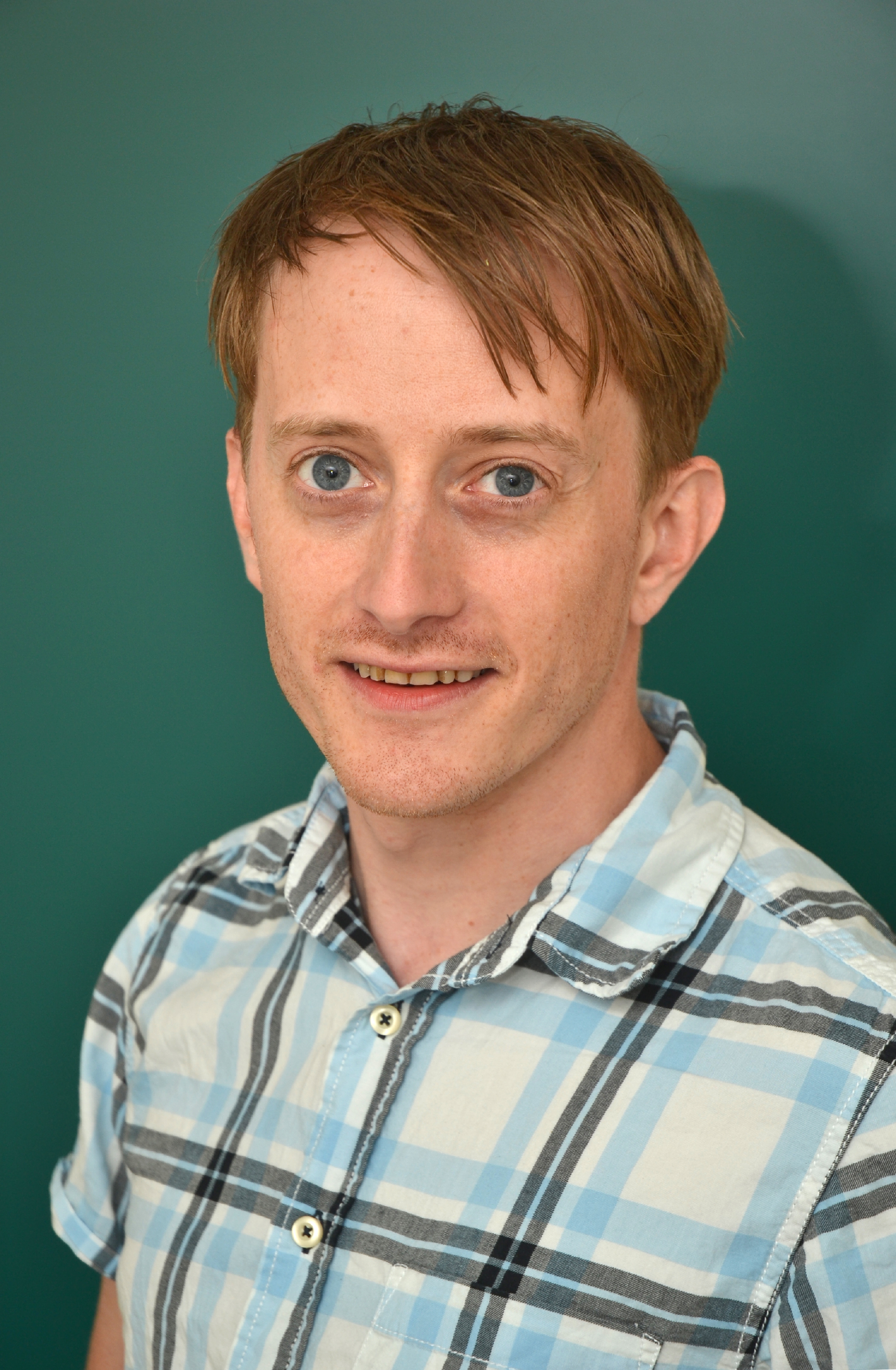}}]{Morten Goodwin} received the B.Sc. and M.Sc. degrees from the University of Agder, Norway, in 2003 and 2005, respectively, and the Ph.D. degree from Aalborg University Department of Computer Science, Denmark, in 2011, on applying machine learning algorithms on eGovernment indicators which are difficult to measure automatically.
He is a Professor with the Department of ICT, the University of Agder, deputy director for Centre for Artificial Intelligence Research, a public speaker, and an active researcher.
His main research interests include machine learning, including swarm intelligence, deep learning, and adaptive learning in medicine, games, and chatbots. He has more than 100 peer reviews of scientific publications. He has supervised more than 110 student projects, including Master and Ph. D. theses within these topics, and more than 90 popular science public speaking events, mostly in Artificial Intelligence.
\end{IEEEbiography}

\begin{landscape}
\begin{table}
\caption{State transitions per scenarios, in the noise-free case, where $P(y=1 | X=1) = 1$, and $P(y=0 | X=0) = 1$;\\ $P(X=1)=c$; $c \in (0,1)$.}
\label{table:noise-free}
\begin{tabular}[]{|l|l|l|l|l|l|}
\hline
{\textbf{$(\mathrm{TA}_1, \mathrm{TA}_2)$}}&
{Scenario index}&
$(X, y)$&
$C(X)$&
{\textbf{State transitions for $\mathrm{TA}_1$}:}&
{\textbf{State transitions for $\mathrm{TA}_2$}:}\\

\hline
\hline

\multirow{2}{*}
{$(E(x), E(\neg x))$}
&
Scenario 1
&
$(1,1)$
&
1
&
\begin{minipage}{0.45\textwidth}
\begin{tikzpicture}[node distance = .35cm, font=\Huge]
    \tikzstyle{every node}=[scale=0.35]
    
    \node[state] (E) at (1,1) {};
    \node[state] (F) at (2,1) {};
    \node[state] (G) at (3,1) {};
    \node[state] (H) at (4,1) {};
    
    \node[thick] at (0,1) {$P$};
    \node[thick] at (1.5,2) {$I$};
    \node[thick] at (3.5,2) {$E$};
    
    \draw[dotted, thick] (2.5,0.5) -- (2.5,1.5);
    
    \draw[every loop]
    (H) edge[bend right] node [scale=1.2, above=0.1 of G] {$\frac{(s-1)c}{s}$} (G)
    (G) edge[bend right] node {} (F);

\end{tikzpicture}
\end{minipage}
&
\begin{minipage}{0.45\textwidth}
\begin{tikzpicture}[node distance = .35cm, font=\Huge]
    \tikzstyle{every node}=[scale=0.35]
    
    \node[state] (E) at (1,1) {};
    \node[state] (F) at (2,1) {};
    \node[state] (G) at (3,1) {};
    \node[state] (H) at (4,1) {};
    
    \node[thick] at (0,1) {$R$};
    \node[thick] at (1.5,2) {$I$};
    \node[thick] at (3.5,2) {$E$};
    
    \draw[dotted, thick] (2.5,0.5) -- (2.5,1.5);
    
    \draw[every loop]
    (G) edge[bend left] node {} (H)
    (H) edge[loop right] node [scale=1.2, below=0.1 of H] {$\frac{c}{s}$} (H);

\end{tikzpicture}
\end{minipage}
\\
\cline{2-6}

&
Scenario 2
&
$(0,0)$
&
1
&
\begin{minipage}{0.45\textwidth}
\begin{tikzpicture}[node distance = .35cm, font=\Huge]
    \tikzstyle{every node}=[scale=0.35]
    
    \node[state] (E) at (1,1) {};
    \node[state] (F) at (2,1) {};
    \node[state] (G) at (3,1) {};
    \node[state] (H) at (4,1) {};
    
    \node[thick] at (0,1) {$P$};
    \node[thick] at (1.5,2) {$I$};
    \node[thick] at (3.5,2) {$E$};
    
    \draw[dotted, thick] (2.5,0.5) -- (2.5,1.5);
    
    \draw[every loop]
    (H) edge[bend right] node [scale=1.2, above=0.1 of G] {$1-c$} (G)
    (G) edge[bend right] node {} (F);

\end{tikzpicture}
\end{minipage}
&
\begin{minipage}{0.45\textwidth}
No change.
\end{minipage}
\\
\hline

\multirow{2}{*}
{$(E(x), I(\neg x))$}
&
Scenario 3
&
$(1,1)$
&
0
&
\begin{minipage}{0.45\textwidth}
\begin{tikzpicture}[node distance = .35cm, font=\Huge]
    \tikzstyle{every node}=[scale=0.35]
    
    \node[state] (E) at (1,1) {};
    \node[state] (F) at (2,1) {};
    \node[state] (G) at (3,1) {};
    \node[state] (H) at (4,1) {};
  
    \node[thick] at (0,1) {$R$};
    \node[thick] at (1.5,2) {$I$};
    \node[thick] at (3.5,2) {$E$};
    
    \draw[dotted, thick] (2.5,0.5) -- (2.5,1.5);
    
    \draw[every loop]
    (G) edge[bend left] node {} (H)
    (H) edge[loop right] node [scale=1.2, below=0.1 of H] {$\frac{c}{s}$} (H);

\end{tikzpicture}
\end{minipage}
&
\begin{minipage}{0.45\textwidth}
\begin{tikzpicture}[node distance = .35cm, font=\Huge]
    \tikzstyle{every node}=[scale=0.35]
    
    \node[state] (E) at (1,1) {};
    \node[state] (F) at (2,1) {};
    \node[state] (G) at (3,1) {};
    \node[state] (H) at (4,1) {};
  
    \node[thick] at (0,1) {$P$};
    \node[thick] at (1.5,2) {$I$};
    \node[thick] at (3.5,2) {$E$};
    
    \draw[dotted, thick] (2.5,0.5) -- (2.5,1.5);
    
    \draw[every loop]
    (E) edge[bend left] node [scale=1.2, above=0.1 of G] {$\frac{c}{s}$} (F)
    (F) edge[bend left] node {} (G);

\end{tikzpicture}
\end{minipage}
\\
\cline{2-6}

&
Scenario 4
&
$(0,0)$
&
1
&
\begin{minipage}{0.45\textwidth}
\begin{tikzpicture}[node distance = .35cm, font=\Huge]
    \tikzstyle{every node}=[scale=0.35]
    
    \node[state] (E) at (1,1) {};
    \node[state] (F) at (2,1) {};
    \node[state] (G) at (3,1) {};
    \node[state] (H) at (4,1) {};
    
    \node[thick] at (0,1) {$P$};
    \node[thick] at (1.5,2) {$I$};
    \node[thick] at (3.5,2) {$E$};
    
    \draw[dotted, thick] (2.5,0.5) -- (2.5,1.5);
    
    \draw[every loop]
    (H) edge[bend right] node [scale=1.2, above=0.1 of G] {$1-c$} (G)
    (G) edge[bend right] node {} (F);

\end{tikzpicture}
\end{minipage}
&
\begin{minipage}{0.45\textwidth}
No change.
\end{minipage}
\\
\hline

\multirow{2}{*}
{
$(I(x), E(\neg x))$
}
&
Scenario 5
&
(1,1)
&
1
&
\begin{minipage}{0.45\textwidth}
\begin{tikzpicture}[node distance = .35cm, font=\Huge]
    \tikzstyle{every node}=[scale=0.35]
    
    \node[state] (E) at (1,1) {};
    \node[state] (F) at (2,1) {};
    \node[state] (G) at (3,1) {};
    \node[state] (H) at (4,1) {};
  
    
    \node[thick] at (0,1) {$R$};
    \node[thick] at (1.5,2) {$I$};
    \node[thick] at (3.5,2) {$E$};
    
    \draw[dotted, thick] (2.5,0.5) -- (2.5,1.5);

    \draw[every loop]
    (F) edge[bend left] node {} (E)
    (E) edge[loop left] node [scale=1.2, below=0.1 of E] {$\frac{(s-1)c}{s}$} (E);

\end{tikzpicture}
\end{minipage}
&
\begin{minipage}{0.45\textwidth}
\begin{tikzpicture}[node distance = .35cm, font=\Huge]
    \tikzstyle{every node}=[scale=0.35]
    
    \node[state] (E) at (1,1) {};
    \node[state] (F) at (2,1) {};
    \node[state] (G) at (3,1) {};
    \node[state] (H) at (4,1) {};
    
    \node[thick] at (0,1) {$R$};
    \node[thick] at (1.5,2) {$I$};
    \node[thick] at (3.5,2) {$E$};
    
    \draw[dotted, thick] (2.5,0.5) -- (2.5,1.5);
    
    \draw[every loop]
    (G) edge[bend left] node {} (H)
    (H) edge[loop right] node [scale=1.2, below=0.1 of H] {$\frac{c}{s}$} (H);

\end{tikzpicture}
\end{minipage}
\\
\cline{2-6}

&
Scenario 6
&
(0,0)
&
0
&
\begin{minipage}{0.45\textwidth}
No change.
\end{minipage}
&
\begin{minipage}{0.45\textwidth}
No change.
\end{minipage}
\\
\hline

\multirow{2}{*}
{$(I(x), I(\neg x))$}
&
Scenario 7
&
(1,1)
&
0
&
\begin{minipage}{0.45\textwidth}
\begin{tikzpicture}[node distance = .35cm, font=\Huge]
    \tikzstyle{every node}=[scale=0.35]
    
    \node[state] (E) at (1,1) {};
    \node[state] (F) at (2,1) {};
    \node[state] (G) at (3,1) {};
    \node[state] (H) at (4,1) {};
    
    \node[thick] at (0,1) {$P$};
    \node[thick] at (1.5,2) {$I$};
    \node[thick] at (3.5,2) {$E$};
    
    \draw[dotted, thick] (2.5,0.5) -- (2.5,1.5);
    
    \draw[every loop]
    (E) edge[bend left] node [scale=1.2, above=0.1 of F] {$\frac{c}{s}$} (F)
    (F) edge[bend left] node {} (G);

\end{tikzpicture}
\end{minipage}
&
\begin{minipage}{0.45\textwidth}
\begin{tikzpicture}[node distance = .35cm, font=\Huge]
    \tikzstyle{every node}=[scale=0.35]
    
    \node[state] (E) at (1,1) {};
    \node[state] (F) at (2,1) {};
    \node[state] (G) at (3,1) {};
    \node[state] (H) at (4,1) {};
  
    \node[thick] at (0,1) {$P$};
    \node[thick] at (1.5,2) {$I$};
    \node[thick] at (3.5,2) {$E$};
    
    \draw[dotted, thick] (2.5,0.5) -- (2.5,1.5);
    
    \draw[every loop]
    (E) edge[bend left] node [scale=1.2, above=0.1 of G] {$\frac{c}{s}$} (F)
    (F) edge[bend left] node {} (G);

\end{tikzpicture}
\end{minipage}
\\
\cline{2-6}

&
Scenario 8
&
$(0,0)$
&
0
&
\begin{minipage}{0.45\textwidth}
No change.
\end{minipage}
&
\begin{minipage}{0.45\textwidth}
No change.
\end{minipage}
\\
\hline

\end{tabular}
\end{table}
\end{landscape}
\begin{landscape}
\begin{table}
\caption{State transitions per senarios, part 1, in the case with noise, where $ P(y=1 | X=1) = a$, and $P(y=1 | X=0) = b$; \\$P(X=1)=c$; $a,b,c\in (0,1)$.}
\label{Table:noisycasepart1}
\begin{tabular}[]{|l|l|l|l|l|l|}
\hline
$(\mathrm{TA}_1, \mathrm{TA}_2)$ & Scenario index & $(X,y)$ & $C(X)$ & \textbf{State transitions for $\mathrm{TA}_1$:} & \textbf{State transitions for $\mathrm{TA}_2$:}\\
\hline
\hline

\multirow{4}{*}
{$(E(x), E(\neg x))$}
&
Scenario 1 
&
$(1,1)$
&
$1$
&
\begin{minipage}{0.45\textwidth}
\begin{tikzpicture}[node distance = .35cm, font=\Huge]
    \tikzstyle{every node}=[scale=0.35]
    
    \node[state] (A) at (1,1) {};
    \node[state] (B) at (2,1) {};
    \node[state] (C) at (3,1) {};
    \node[state] (D) at (4,1) {};
    
    \node[thick] at (0,1) {$P$};
    \node[thick] at (1.5,2) {$I$};
    \node[thick] at (3.5,2) {$E$};
    
    \draw[dotted, thick] (2.5,0.5) -- (2.5,1.5);
    
    \draw[every loop]
    (C) edge[bend left] node {} (B)
    (D) edge[bend left] node [scale=1.2, below=0.28 of C] {$\frac{(s-1)ac}{s}$} (C);

\end{tikzpicture}
\end{minipage}
&
\begin{minipage}{0.45\textwidth}
\begin{tikzpicture}[node distance = .35cm, font=\Huge]
    \tikzstyle{every node}=[scale=0.35]
    
    \node[state] (E) at (1,1) {};
    \node[state] (F) at (2,1) {};
    \node[state] (G) at (3,1) {};
    \node[state] (H) at (4,1) {};
 
    \node[thick] at (0,1) {$R$};
    \node[thick] at (1.5,2) {$I$};
    \node[thick] at (3.5,2) {$E$};
    
    \draw[dotted, thick] (2.5,0.5) -- (2.5,1.5);
    
    \draw[every loop]
    (G) edge[bend left] node {} (H)
    (H) edge[loop right] node [scale=1.2, below=0.1 of H] {$\frac{ac}{s}$} (H);

\end{tikzpicture}
\end{minipage}
\\
\cline{2-6}

&
Scenario 2
&
$(1,0)$
&
$0$
&
No change.
&
\begin{minipage}{0.45\textwidth}
\begin{tikzpicture}[node distance = .35cm, font=\Huge]
    \tikzstyle{every node}=[scale=0.35]
    
    \node[state] (A) at (1,1) {};
    \node[state] (B) at (2,1) {};
    \node[state] (C) at (3,1) {};
    \node[state] (D) at (4,1) {};
    
    \node[thick] at (0,1) {$P$};
    \node[thick] at (1.5,2) {$I$};
    \node[thick] at (3.5,2) {$E$};
    
    \draw[dotted, thick] (2.5,0.5) -- (2.5,1.5);
    
    \draw[every loop]
    (C) edge [bend right ] node {} (B)
    (D) edge [bend right ] node [scale=1.2, above=0.15 of C]{$(1-a)c$} (C);
\end{tikzpicture}
\end{minipage}
\\
\cline{2-6}

&
Scenario 3
&
$(0,0)$
&
$1$
&
\begin{minipage}{0.45\textwidth}
\begin{tikzpicture}[node distance = .35cm, font=\Huge]
    \tikzstyle{every node}=[scale=0.35]
    
    \node[state] (A) at (1,1) {};
    \node[state] (B) at (2,1) {};
    \node[state] (C) at (3,1) {};
    \node[state] (D) at (4,1) {};
    
    \node[thick] at (0,1) {$P$};
    \node[thick] at (1.5,2) {$I$};
    \node[thick] at (3.5,2) {$E$};
    
    \draw[dotted, thick] (2.5,0.5) -- (2.5,1.5);
    
    \draw[every loop]
    (C) edge [bend right ] node {} (B)
    (D) edge [bend right ] node [scale=1.2, above=0.15 of C]{$(1-b)(1-c)$} (C);
\end{tikzpicture}
\end{minipage}
&
No change.
\\
\cline{2-6}

&
Scenario 4
&
$(0,1)$
&
$1$
&
\begin{minipage}{0.45\textwidth}
\begin{tikzpicture}[node distance = .35cm, font=\Huge]
    \tikzstyle{every node}=[scale=0.35]
    
    \node[state] (A) at (1,1) {};
    \node[state] (B) at (2,1) {};
    \node[state] (C) at (3,1) {};
    \node[state] (D) at (4,1) {};
    
    \node[thick] at (0,1) {$R$};
    \node[thick] at (1.5,2) {$I$};
    \node[thick] at (3.5,2) {$E$};
    
    \draw[dotted, thick] (2.5,0.5) -- (2.5,1.5);
    
    \draw[every loop]
    (C) edge [bend left] node {} (D)
    (D) edge [loop right] node [scale=1.2, below=0.15 of D] {$\frac{b(1-c)}{s}$} (D);
\end{tikzpicture}
\end{minipage}
&
\begin{minipage}{0.45\textwidth}
\begin{tikzpicture}[node distance = .35cm, font=\Huge]
    \tikzstyle{every node}=[scale=0.35]
    
    \node[state] (E) at (1,1) {};
    \node[state] (F) at (2,1) {};
    \node[state] (G) at (3,1) {};
    \node[state] (H) at (4,1) {};
    
    \node[thick] at (0,1) {$P$};
    \node[thick] at (1.5,2) {$I$};
    \node[thick] at (3.5,2) {$E$} ;
    
    \draw[dotted, thick] (2.5,0.5) -- (2.5,1.5);
    
    \draw[every loop]
    (G) edge [bend right ] node {} (F)
    (H) edge [bend right ] node [scale=1.2, below=0.35 of C] {$\frac{(s-1)b(1-c)}{s}$} (G);

\end{tikzpicture}
\end{minipage}
\\
\hline

\multirow{4}{*}
{$(E(x), I(\neg x))$}
&
Scenario 5
&
$(1,1)$
&
$0$
&
\begin{minipage}{0.45\textwidth}
\begin{tikzpicture}[node distance = .35cm, font=\Huge]
    \tikzstyle{every node}=[scale=0.35]
    
    \node[state] (E) at (1,1) {};
    \node[state] (F) at (2,1) {};
    \node[state] (G) at (3,1) {};
    \node[state] (H) at (4,1) {};
 
    \node[thick] at (0,1) {$R$};
    \node[thick] at (1.5,2) {$I$};
    \node[thick] at (3.5,2) {$E$};
    
    \draw[dotted, thick] (2.5,0.5) -- (2.5,1.5);
    
    \draw[every loop]
    (G) edge[bend left] node {} (H)
    (H) edge[loop right] node [scale=1.2, below=0.1 of H] {$\frac{ac}{s}$} (H);

\end{tikzpicture}
\end{minipage}
&
\begin{minipage}{0.45\textwidth}
\begin{tikzpicture}[node distance = .35cm, font=\Huge]
    \tikzstyle{every node}=[scale=0.35]
    
    \node[state] (A) at (1,1) {};
    \node[state] (B) at (2,1) {};
    \node[state] (C) at (3,1) {};
    \node[state] (D) at (4,1) {};
    
    \node[thick] at (0,1) {$P$};
    \node[thick] at (1.5,2) {$I$};
    \node[thick] at (3.5,2) {$E$};
    
    \draw[dotted, thick] (2.5,0.5) -- (2.5,1.5);
    
    \draw[every loop]
    (A) edge[bend left] node [scale=1.2, above=0.1 of B] {$\frac{ac}{s}$} (B)
    (B) edge[bend left] node {} (C);

\end{tikzpicture}
\end{minipage}
\\
\cline{2-6}

&
Scenario 6
&
$(1,0)$
&
$0$
&
No change.
&
No change.
\\
\cline{2-6}

&
Scenario 7
&
$(0,0)$
&
$1$
&
\begin{minipage}{0.45\textwidth}
\begin{tikzpicture}[node distance = .35cm, font=\Huge]
    \tikzstyle{every node}=[scale=0.35]
    
    \node[state] (A) at (1,1) {};
    \node[state] (B) at (2,1) {};
    \node[state] (C) at (3,1) {};
    \node[state] (D) at (4,1) {};
    
    \node[thick] at (0,1) {$P$};
    \node[thick] at (1.5,2) {$I$};
    \node[thick] at (3.5,2) {$E$};
    
    \draw[dotted, thick] (2.5,0.5) -- (2.5,1.5);
    
    \draw[every loop]
    (C) edge [bend right ] node {} (B)
    (D) edge [bend right ] node [scale=1.2, above=0.15 of C]{$(1-b)(1-c)$} (C);
\end{tikzpicture}
\end{minipage}
&
No change.
\\
\cline{2-6}

&
Scenario 8
&
$(0,1)$
&
$1$
&
\begin{minipage}{0.45\textwidth}
\begin{tikzpicture}[node distance = .35cm, font=\Huge]
    \tikzstyle{every node}=[scale=0.35]
    
    \node[state] (A) at (1,1) {};
    \node[state] (B) at (2,1) {};
    \node[state] (C) at (3,1) {};
    \node[state] (D) at (4,1) {};
    
    \node[thick] at (0,1) {$R$};
    \node[thick] at (1.5,2) {$I$};
    \node[thick] at (3.5,2) {$E$};
    
    \draw[dotted, thick] (2.5,0.5) -- (2.5,1.5);
    
    \draw[every loop]
    (C) edge [bend left] node {} (D)
    (D) edge [loop right] node [scale=1.2, below=0.15 of D] {$\frac{b(1-c)}{s}$} (D);
\end{tikzpicture}
\end{minipage}
&
\begin{minipage}{0.45\textwidth}
\begin{tikzpicture}[node distance = .35cm, font=\Huge]
    \tikzstyle{every node}=[scale=0.35]
    
    \node[state] (E) at (1,1) {};
    \node[state] (F) at (2,1) {};
    \node[state] (G) at (3,1) {};
    \node[state] (H) at (4,1) {};

    \node[thick] at (0,1) {$R$};
    \node[thick] at (1.5,2) {$I$};
    \node[thick] at (3.5,2) {$E$};
    
    \draw[dotted, thick] (2.5,0.5) -- (2.5,1.5);
    
    \draw[every loop]
    (F) edge[bend left] node {} (E)
    (E) edge[loop left] node [scale=1.2, below=0.1 of E] {$\frac{(s-1)b(1-c)}{s}$} (E);

\end{tikzpicture}
\end{minipage}
\\
\hline

\end{tabular}
\end{table}
\end{landscape}
\begin{landscape}
\begin{table}
\caption{State transitions per scenarios, part 2, in the case with noise, where $ P(y=1 | X=1) = a$, and $P(y=1 | X=0) = b$; \\$P(X=1)=c$; $a,b,c\in (0,1)$.}
\label{Table:noisycasepart2}
\begin{tabular}[]{|l|l|l|l|l|l|}
\hline
$(\mathrm{TA}_1, \mathrm{TA}_2)$ & Scenario index &$(X,y)$ & $C(X)$ & \textbf{State transitions for $\mathrm{TA}_1$:} & \textbf{State transitions for $\mathrm{TA}_2$:}\\
\hline
\hline

\multirow{4}{*}
{$(I(x), E(\neg x))$}
&
Scenario 9
&
$(1,1)$
&
$1$
&
\begin{minipage}{0.45\textwidth}
\begin{tikzpicture}[node distance = .35cm, font=\Huge]
    \tikzstyle{every node}=[scale=0.35]
    \node[state] (E) at (1,1) {};
    \node[state] (F) at (2,1) {};
    \node[state] (G) at (3,1) {};
    \node[state] (H) at (4,1) {};
  
    \node[thick] at (0,1) {$R$};
    \node[thick] at (1.5,2) {$I$};
    \node[thick] at (3.5,2) {$E$};
    
    \draw[dotted, thick] (2.5,0.5) -- (2.5,1.5);
    
    \draw[every loop]
    (F) edge[bend left] node {} (E)
    (E) edge[loop left] node [scale=1.2, below=0.1 of E] {$\frac{(s-1)ac}{s}$} (E);

\end{tikzpicture}
\end{minipage}
&
\begin{minipage}{0.45\textwidth}
\begin{tikzpicture}[node distance = .35cm, font=\Huge]
    \tikzstyle{every node}=[scale=0.35]
    
    \node[state] (E) at (1,1) {};
    \node[state] (F) at (2,1) {};
    \node[state] (G) at (3,1) {};
    \node[state] (H) at (4,1) {};
 
    \node[thick] at (0,1) {$R$};
    \node[thick] at (1.5,2) {$I$};
    \node[thick] at (3.5,2) {$E$};
    
    \draw[dotted, thick] (2.5,0.5) -- (2.5,1.5);
    
    \draw[every loop]
    (G) edge[bend left] node {} (H)
    (H) edge[loop right] node [scale=1.2, below=0.1 of H] {$\frac{ac}{s}$} (H);

\end{tikzpicture}
\end{minipage}
\\
\cline{2-6}

&
Scenario 10
&
$(1,0)$
&
$1$
&
No change.
&
\begin{minipage}{0.45\textwidth}
\begin{tikzpicture}[node distance = .35cm, font=\Huge]
    \tikzstyle{every node}=[scale=0.35]
    
    \node[state] (A) at (1,1) {};
    \node[state] (B) at (2,1) {};
    \node[state] (C) at (3,1) {};
    \node[state] (D) at (4,1) {};
    
    \node[thick] at (0,1) {$P$};
    \node[thick] at (1.5,2) {$I$};
    \node[thick] at (3.5,2) {$E$};
    
    \draw[dotted, thick] (2.5,0.5) -- (2.5,1.5);
    
    \draw[every loop]
    (C) edge [bend right ] node {} (B)
    (D) edge [bend right ] node [scale=1.2, above=0.15 of C]{$(1-a)c$} (C);
\end{tikzpicture}
\end{minipage}
\\
\cline{2-6}

&
Scenario 11
&
$(0,0)$
&
$0$
&
No change.
&
No change.
\\
\cline{2-6}

&
Scenario 12
&
$(0,1)$
&
$0$
&
\begin{minipage}{0.45\textwidth}
\begin{tikzpicture}[node distance = .35cm, font=\Huge]
    \tikzstyle{every node}=[scale=0.35]
    
    \node[state] (A) at (1,1) {};
    \node[state] (B) at (2,1) {};
    \node[state] (C) at (3,1) {};
    \node[state] (D) at (4,1) {};
    
    \node[thick] at (0,1) {$P$};
    \node[thick] at (1.5,2) {$I$};
    \node[thick] at (3.5,2) {$E$};
    
    \draw[dotted, thick] (2.5,0.5) -- (2.5,1.5);
    
    \draw[every loop]
    (A) edge[bend left] node [scale=1.2, above=0.1 of B] {$\frac{b(1-c)}{s}$} (B)
    (B) edge[bend left] node{} (C);

\end{tikzpicture}
\end{minipage}
&
\begin{minipage}{0.45\textwidth}
\begin{tikzpicture}[node distance = .35cm, font=\Huge]
    \tikzstyle{every node}=[scale=0.35]
    
    \node[state] (E) at (1,1) {};
    \node[state] (F) at (2,1) {};
    \node[state] (G) at (3,1) {};
    \node[state] (H) at (4,1) {};
    
    \node[thick] at (0,1) {$R$};
    \node[thick] at (1.5,2) {$I$};
    \node[thick] at (3.5,2) {$E$};
    
    \draw[dotted, thick] (2.5,0.5) -- (2.5,1.5);
    
    \draw[every loop]
    (G) edge[bend left] node {} (H)
    (H) edge[loop right] node [scale=1.2, below=0.1 of H] {$\frac{b(1-c)}{s}$} (H);

\end{tikzpicture}
\end{minipage}
\\
\hline

\multirow{4}{*}
{$(I(x), I(\neg x))$}
&
Scenario 13
&
$(1,1)$
&
$0$
&
\begin{minipage}{0.45\textwidth}
\begin{tikzpicture}[node distance = .35cm, font=\Huge]
    \tikzstyle{every node}=[scale=0.35]
    
    \node[state] (A) at (1,1) {};
    \node[state] (B) at (2,1) {};
    \node[state] (C) at (3,1) {};
    \node[state] (D) at (4,1) {};
    
    \node[thick] at (0,1) {$P$};
    \node[thick] at (1.5,2) {$I$};
    \node[thick] at (3.5,2) {$E$};
    
    \draw[dotted, thick] (2.5,0.5) -- (2.5,1.5);
    
    \draw[every loop]
    (A) edge[bend left] node [scale=1.2, above=0.1 of B] {$\frac{ac}{s}$} (B)
    (B) edge[bend left] node {} (C);

\end{tikzpicture}
\end{minipage}
&
\begin{minipage}{0.45\textwidth}
\begin{tikzpicture}[node distance = .35cm, font=\Huge]
    \tikzstyle{every node}=[scale=0.35]
    
    \node[state] (A) at (1,1) {};
    \node[state] (B) at (2,1) {};
    \node[state] (C) at (3,1) {};
    \node[state] (D) at (4,1) {};
    
    \node[thick] at (0,1) {$P$};
    \node[thick] at (1.5,2) {$I$};
    \node[thick] at (3.5,2) {$E$};
    
    \draw[dotted, thick] (2.5,0.5) -- (2.5,1.5);
    
    \draw[every loop]
    (A) edge[bend left] node [scale=1.2, above=0.1 of B] {$\frac{ac}{s}$} (B)
    (B) edge[bend left] node {} (C);

\end{tikzpicture}
\end{minipage}
\\
\cline{2-6}

&
Scenario 14
&
$(1,0)$
&
$0$
&
No change.
&
No change.
\\
\cline{2-6}

&
Scenario 15
&
$(0,0)$
&
$0$
&
No change.
&
No change.
\\
\cline{2-6}

&
Scenario 16
&
$(0,1)$
&
$0$
&
\begin{minipage}{0.45\textwidth}
\begin{tikzpicture}[node distance = .35cm, font=\Huge]
    \tikzstyle{every node}=[scale=0.35]
    
    \node[state] (A) at (1,1) {};
    \node[state] (B) at (2,1) {};
    \node[state] (C) at (3,1) {};
    \node[state] (D) at (4,1) {};
    
    \node[thick] at (0,1) {$P$};
    \node[thick] at (1.5,2) {$I$};
    \node[thick] at (3.5,2) {$E$};
    
    \draw[dotted, thick] (2.5,0.5) -- (2.5,1.5);
    
    \draw[every loop]
    (A) edge[bend left] node [scale=1.2, above=0.1 of B] {$\frac{b(1-c)}{s}$} (B)
    (B) edge[bend left] node{} (C);

\end{tikzpicture}
\end{minipage}
&
\begin{minipage}{0.45\textwidth}
\begin{tikzpicture}[node distance = .35cm, font=\Huge]
    \tikzstyle{every node}=[scale=0.35]
    
    \node[state] (A) at (1,1) {};
    \node[state] (B) at (2,1) {};
    \node[state] (C) at (3,1) {};
    \node[state] (D) at (4,1) {};
    
    \node[thick] at (0,1) {$P$};
    \node[thick] at (1.5,2) {$I$};
    \node[thick] at (3.5,2) {$E$};
    
    \draw[dotted, thick] (2.5,0.5) -- (2.5,1.5);
    
    \draw[every loop]
    (A) edge[bend left] node [scale=1.2, above=0.1 of B] {$\frac{b(1-c)}{s}$} (B)
    (B) edge[bend left] node  {} (C);

\end{tikzpicture}
\end{minipage}
\\
\hline

\end{tabular}
\end{table}
\end{landscape}


%

\ifCLASSOPTIONcaptionsoff
  \newpage
\fi

\end{document}